\documentclass{article}
\pdfoutput=1

\usepackage[preprint,nonatbib]{neurips_2020}

\usepackage[utf8]{inputenc} %
\usepackage[T1]{fontenc}    %
\usepackage{array,makecell}
\usepackage{amsfonts}       %
\usepackage{amsmath}
\usepackage{amsthm}
\usepackage{hyperref}       %
\usepackage{cleveref}       %
\usepackage{enumitem}
\usepackage{url}            %
\usepackage{booktabs}       %
\usepackage{pdflscape}
\usepackage{mathtools}
\usepackage{microtype}      %
\usepackage{multirow}
\usepackage{nicefrac}       %
\usepackage{graphicx}
\usepackage{dsfont}
\usepackage{subcaption}

\usepackage[normalem]{ulem}
\usepackage{xcolor,colortbl}
\usepackage{xspace}
\usepackage{comment}
\usepackage[rightcaption]{sidecap}

\newcommand{\cit}[2][l]{\shortstack[#1]{#2 \\ {}}}
\newcommand{\civ}[4][r]{\shortstack[#1]{#2 \\ \scalebox{0.5}{[#3--#4]}}}
\newcommand{\cih}[3]{#1 \scalebox{0.6}{[#2--#3]}}

\DeclareMathOperator*{\argmax}{arg\,max}
\DeclareMathOperator*{\argmin}{arg\,min}

\DeclarePairedDelimiterX{\divx}[2]{(}{)}{#1\;\delimsize\|\;#2}
\newcommand{\divkl}{D_{\text{KL}}\divx}

\newcommand{\EPN}{\mathrm{EPN}}

\newcommand{\JFT}{{JFT}\xspace}

\definecolor{naturalcolor}{rgb}{0.835,0.369,0.0}
\definecolor{specializedcolor}{rgb}{0.800,0.475,0.655}
\definecolor{structuredcolor}{rgb}{0.941,0.894,0.259}
\newcommand{\naturalsym}{{\protect\scalebox{1.5}{\color{naturalcolor!50}$\bullet$}}}
\newcommand{\specializedsym}{{\protect\scalebox{1.5}{\color{specializedcolor!50}$\bullet$}}}
\newcommand{\structuredsym}{{\protect\scalebox{1.5}{\color{structuredcolor!50}$\bullet$}}}

\title{Scalable Transfer Learning with Expert Models}

\author{%
  Joan Puigcerver\thanks{Equal contribution. Order decided by a coin toss.} \\
  Google Research \\
  \And 
  Carlos Riquelme\footnote[1]{} \\
  Google Research \\  
  \And 
  Basil Mustafa \\
  Google Research \\  
  \And 
  Cedric Renggli\thanks{Work done while interning at Google Research.} \\
  ETH Zurich\\
  \And 
  Andr\'e Susano Pinto \\
  Google Research \\  
  \And 
  Sylvain Gelly \\
  Google Research \\  
  \And 
  Daniel Keysers \\
  Google Research \\  
  \And 
  Neil Houlsby\\
  Google Research \\  
}

\begin{document}

\maketitle

\begin{abstract}
Transfer of pre-trained representations can improve sample efficiency and reduce computational requirements for new tasks. However, representations used for transfer are usually generic, and are not tailored to a particular distribution of downstream tasks. We explore the use of expert representations for transfer with a simple, yet effective, strategy. We train a diverse set of experts by exploiting existing label structures, and use cheap-to-compute performance proxies to select the relevant expert for each target task. 
This strategy scales the process of transferring to new tasks, since it does not revisit the pre-training data during transfer.
Accordingly, it requires little extra compute per target task, and results in a speed-up of 2--3 orders of magnitude compared to competing approaches.
Further, we provide an adapter-based architecture able to compress many experts into a single model. We evaluate our approach on two different data sources and demonstrate that it outperforms baselines on over 20 diverse vision tasks in both cases.
\end{abstract}
\section{Introduction}

Deep learning has been successful on many computer vision tasks.
Unfortunately, this success often requires a large amount of per-task data and compute.
To scale deep learning to new vision tasks, practitioners often turn to transfer learning.
Transfer learning involves re-using models trained on a large \emph{source} task, and tuning them on the \emph{target} task.
This can improve both convergence rates \cite{ben2007analysis, ben2010theory, blitzer2008learning,
du2017hypothesis,
kuzborskij2013stability,
mansour2009domain,
mansour2009domainsources} and empirical performance \cite{dai2007boosting, donahue2014decaf, oquab2014learning, sharif2014cnn, tan2018survey}.
Transfer learning reduces \emph{per-task} data or compute requirements, given a large one-off pre-training cost.
In practice, this one-off down payment may not be made by the practitioner, since pre-trained networks are made available through platforms like PyTorch Hub \cite{website:pythorch-hub}, TensorFlow Hub \cite{website:tfhub}, and others.
For instance, ImageNet pre-training is popular since it is freely
available and works well for many tasks 
\cite{donahue2014decaf, oquab2014learning, sharif2014cnn}.

In contrast to generic homogeneous models (e.g.\ most pre-trained ImageNet networks), Mixture of Experts (MoE) include 
multiple heterogeneous sub-models (``experts'') that specialize to sub-problems of the full task.
MoEs have been studied for decades \cite{eigen2013learning, jacobs1993learning}, and have also been successful 
in deep learning \cite{shazeer2017outrageously}.
Yet, the application of experts for deep transfer learning has been less explored.
We study visual transfer with experts, and present a simple, scalable, yet effective strategy.

Transfer of specialist models has been studied before.
However, previous approaches (e.g.~\cite{ngiam2018domain,dvornik2020selecting,yan2020neural}) are limited in their scalability and task diversity.
They either require expensive re-training on the source dataset for every target task,
or operate at a small scale where all experts can be applied simultaneously.
Further, most of them are tested only on a limited suite of natural single-object classification tasks. %
We lift these constraints, and present a practical approach that scales to hundreds of large experts, %
while requiring relatively little compute \emph{per target task}. 

\begin{figure}[tb]
\centering
  \includegraphics[width=\textwidth]{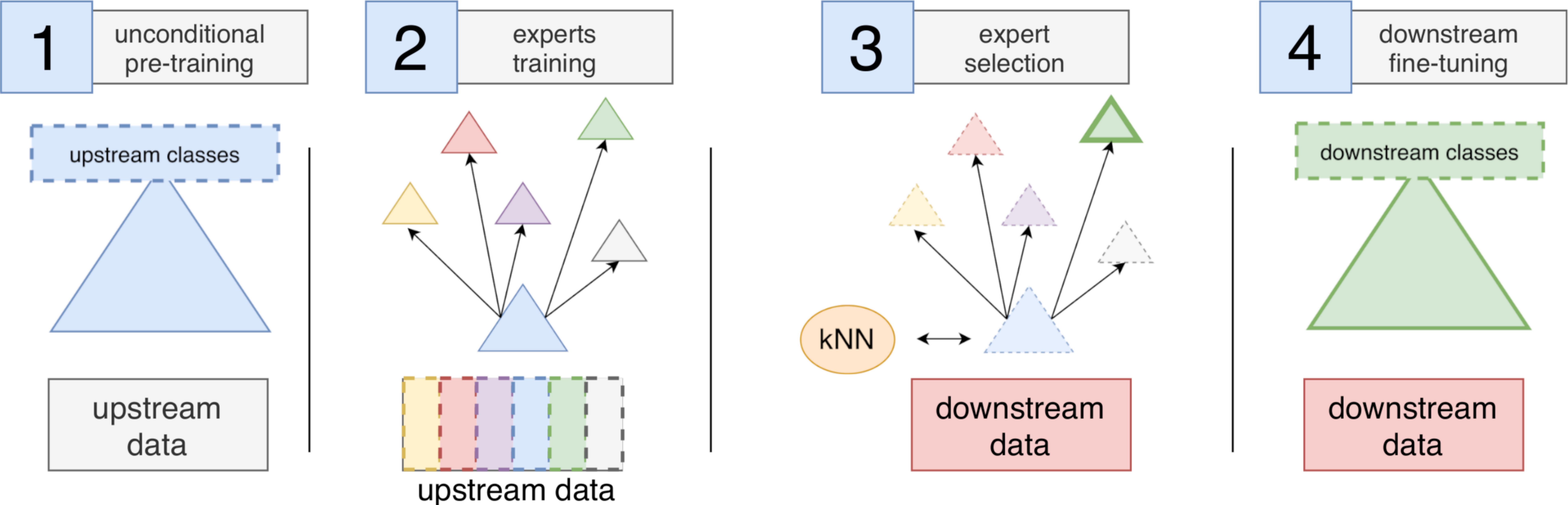}
  \caption{Transfer Learning with Per-Task Routing of Experts.
  \textbf{Step 1.} A single baseline model \textbf{B} is trained on the entire upstream dataset.
  \textbf{Step 2.} The upstream data is divided in semantic subsets (possibly overlapping). One expert is trained on each subset 
  using the weights from \textbf{B} as initialization.
  \textbf{Step 3.} Given a new downstream task $\mathbf{D}_T = (X_T, Y_T)$, we compute the image representations $M_e(X_T)$ from each expert $e$. %
  We use kNN to compute the accuracy on the supervised problem $\mathbf{D}_{T, e} = (M_e(X_T), Y_T)$, and select the expert $e^*$ with highest accuracy.
  \textbf{Step 4.} %
  We add a new head to $e^*$
  and fine-tune its whole network with the downstream data, leading to the final model.
  \label{fig:algorithm_overview}}
  \vspace{-1em}
\end{figure}

Our strategy consists of four stages (\cref{fig:algorithm_overview}).
(1) \emph{Unconditional pre-training}. A single baseline model is trained on the entire upstream data.
(2) \emph{Experts training}.
Multiple experts are pre-trained by exploiting the label hierarchy present in many large-scale image datasets, such as ImageNet and JFT.
In addition to entire expert networks, we explore residual adapters
that allow all of the expertise to be packed into a single model that can be loaded into memory.
These two stages may be expensive, but are done only once.
(3) \emph{Expert selection}.
Applying all experts to each task does not scale well; some sort of sparsification is required.
We focus on inexpensive model selection that can be applied to hundreds or thousands of experts.
(4) \emph{Downstream fine-tuning}.
We take the output of the model selection phase and tune it on the target task.
Importantly, this phase does not require revisiting the source dataset, which may be unavailable or expensive to train on.

\begin{figure}[htb]
\centering
\includegraphics[width=\textwidth]{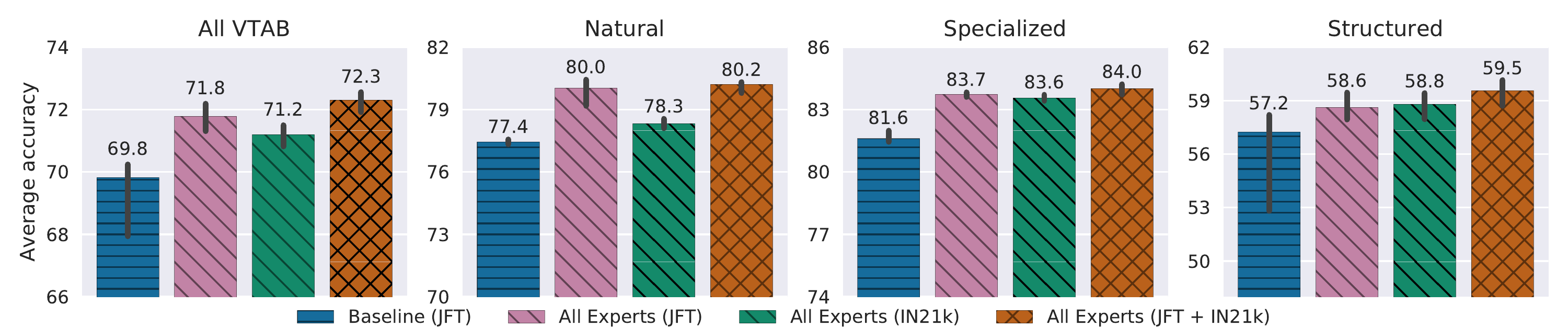}
\caption{Summary of results on the VTAB-1k benchmark, combining experts 
with different architectures trained on two different data sources (JFT, ImageNet21k). In each of the 19 datasets, we use the median accuracy over
30 runs.
The average of the accuracies in each group is shown, as well as (percentile) bootstrap confidence intervals at the 95\% level.
\label{fig:vtab_test_all_experts}}
\vspace{-1em}
\end{figure}

We show that this approach yields remarkably strong performance on many diverse tasks.
We evaluate not only on classic vision tasks (Oxford Pets \cite{parkhi12a}, Stanford Cars \cite{KrauseStarkDengFei-Fei_3DRR2013}, etc.), but also on the diverse VTAB benchmark of 19 tasks~\cite{zhai2019visual}.
Our contributions can be summarized as follows.
\begin{itemize}[leftmargin=*]
     \item We propose a 
     transfer learning algorithm with a large number of experts based on per-task routing via nearest neighbors selection.
     Once we have amortized the pre-training cost, this algorithm requires little compute \emph{per target task},
     achieving an speed-up of $500\times$--$1000\times$ compared to competing strategies.
     Also, it can be easily replicated with any large upstream multilabel dataset.
     \item %
     We achieve a mean accuracy improvement of %
     3.6\% over the
     state-of-the-art performance on 19 VTAB datasets using ResNet50 networks.
     Our algorithm offers improvements on every group of tasks: natural, specialized, and structured. \Cref{fig:vtab_test_all_experts} summarizes
     these results.
     \item We explore using sub-networks as experts via residual adapters, allowing all experts to be packed into a single model.
     Surprisingly these perform almost as well as their full-network counterparts.
\end{itemize}

\section{Related Work}
\label{sec:literature}
\textbf{Transfer Learning.}
Tasks with little training data can benefit from other larger datasets, often from a similar domain.
Transfer learning concerns the link between the source and target dataset~\cite{pan2009survey, weiss2016survey, tan2018survey, wang2018theoretical}.
One family of methods creates a single training dataset, where source instances are 
re-weighted according to their relevance
\cite{dai2007boosting, pardoe2010boosting, wan2011bi, xu2017unified}. 
Alternative approaches learn 
a suitable projection of the source and target data to find useful common features 
reducing domain discrepancy 
\cite{long2015learning, long2017deep, pan2010domain, tzeng2014deep, yosinski2014transferable}. 
Finally, a popular method consists of fine-tuning a model that was 
pre-trained on the source data \cite{donahue2014decaf, oquab2014learning, sharif2014cnn}.
Some transfer learning algorithms condition the initial source
model on the target dataset itself \cite{ngiam2018domain,xie2019self,yalniz2019billion}, while
others (like ours) are agnostic about the downstream task when
the initial model is trained on the source data 
\cite{kolesnikov2019large}. We offer an in-depth comparison with
\cite{ngiam2018domain} in \cref{subsec:adaptive_transfer_comparison}.
In the context of few-shot learning, where out-of-the-box fine-tuning may not work, 
generic representations are sometimes frozen, and simple feature selection 
\cite{dvornik2020selecting} or model training \cite{chen2019closer} techniques
are applied on top.
Instead of relying on fixed universal representations, 
\cite{rebuffi2017learning, rebuffi2018efficient, rosenfeld2018incremental} 
use small additional modules, or adapters, that incorporate knowledge from several visual domains. Our work also explores this idea.

\textbf{Multi-task Learning.} MTL tries to leverage the common aspects of several learning tasks \cite{caruana1997multitask}.
A prominent approach uses explicit parameter sharing; for instance, by means of common low-level layers leading to different heads. Among others, this has been successfully applied to vision \cite{zhang2014facial}, language \cite{liu2015representation}, and reinforcement learning \cite{fedus2019hyperbolic} tasks.
In addition, a variety of ways to combine task-specific representations have arisen, such as cross-stitch networks \cite{misra2016cross}, or lateral connections~\cite{rusu2016progressive}. 
A different family of methods impose joint constraints on the --possibly different-- models corresponding to each task.
We can combine the learning problems
via regularization and shared sparsity patterns \cite{argyriou2007multi, lounici2009taking}, or by imposing some prior knowledge regarding the task structure \cite{evgeniou2005learning, jacob2009clustered, kim2012tree}.

\section{The Transfer Learning Framework}

In this section, we describe our transfer learning setup of interest.
The high-level goal is to train strong models for arbitrary downstream tasks, possibly under severe data and compute limitations.
To do so efficiently, one can offload computation 
to a previous upstream phase which is executed \emph{a priori}, without knowing the downstream tasks in advance.
Accordingly, the upstream model should not depend on any specific target data. 
We are mostly interested in the \emph{low data} regime where downstream tasks contain few datapoints.
These restrictions have a practical motivation: we would like to build and deploy universal representations that 
are easily transferred to a wide range of downstream settings.
Any transfer algorithm must implement the following three stages.

\textbf{Upstream Training.}
Given the upstream data $\mathbf{D}_U$, the algorithm first outputs a source model $\mathbf{M}$.
The goal is to provide useful initial representations for various new tasks.
This stage could actually produce a family of models $\{\mathbf{M}_e\}$ rather than a single one.
These models might not be disjoint, and could share parameters. 
The upstream learning problems are auxiliary; accordingly, $\mathbf{D}_U$ 
could include a diverse set of classification, regression, or even synthetic learning instances.

\textbf{Model Selection.}
When a new downstream task is given, a selection algorithm is applied to choose the upstream model(s) to transfer, possibly depending on the downstream 
data. This phase should be computationally cheap; thus, the upstream data is no longer available.
Sometimes, there is no choice to make (say, with a single ImageNet representation).
Alternatively,
in models with a complex structure, one may choose which parts, routes, or modules to keep in a  data-dependent fashion.

\textbf{Downstream Training.}
The final stage fine-tunes the selected model using the downstream data, either fully or partially.
For neural nets, a new head is added as the output classes are task-specific.

Our overall algorithm is depicted in \cref{fig:algorithm_overview}. We give details about each step in the
following sections.
\section{Upstream Training}
\label{sec:upstream}

In this section, we introduce the two specific architectures we explored for the expert modules, and we explain some key design choices we made for training our experts.

\subsection{Expert Architectures}

Our experts should provide feature extractions that are a good
starting point to learn future tasks related to the expert's
upstream training data.
We explore two different model architectures to train such experts.
As an obvious choice, we first look at Residual Networks \cite{he2016deep}, or ResNets.
These are powerful models; however, storing and deploying many of 
them can be challenging.
As an alternative, we also develop more compact adapter modules that can 
all be assembled in a single architecture.
Also, their individual size can be easily customized to meet memory and 
computational constraints, which makes them an ideal candidate for 
combining multiple experts in a single model, when needed.
We informally refer to these as \emph{full} and \emph{adapter} modules or experts, respectively.

\textbf{Full ResNet Modules.} As a base architecture for full experts we use ResNets.
In particular, all of our experiments focus on the ResNet50-v2 architecture (R50) \cite{He2016Identity}, which sequentially stacks a root block and 4 blocks with (3, 4, 6, 3) residual units.
The initial step in every experiment consists of training a baseline model $\mathbf{B}$ on the whole upstream data (see stage 1 in \cref{fig:algorithm_overview}).
This baseline is subsequently fine-tuned by both full and adapter experts, but in different ways.
A full expert trained on a slice of data is simply the baseline $\mathbf{B}$ fine-tuned on that data.
The head will later be discarded for transfer.
This approach requires as many R50s as there are experts.

\textbf{Adapter Modules.}
Residual adapters were proposed to adapt a neural network to a particular downstream task without needing to fine-tuning the entire network \cite{rebuffi2017learning}.
Originally, they were $1\times1$ convolutional 
layers  that are placed after each $3\times3$ convolution, with a residual connection.
Instead, we use them to adapt the baseline architecture 
to slices of the \emph{upstream data}. 
Also, we do not place them after each 
$3\times3$ convolution, but before each of the R50's blocks. 
Finally, our adapters have a bottleneck and are non-linear, 
as in \cite{houlsby2019parameter}. We insert several in parallel into the backbone $\mathbf{B}$.
When creating an expert, only the adapters are tuned and the backbone weights are frozen.

\begin{figure}[tb]
\centering
\subcaptionbox{\label{fig:resnet_with_adapters}}{%
\includegraphics[width=0.85\textwidth]{images/Horizontal_Adapter_Diagram_Block_Notation.pdf}}
~~~~~~
\subcaptionbox{\label{fig:adapter_arch}}{%
\includegraphics[width=0.1\textwidth]{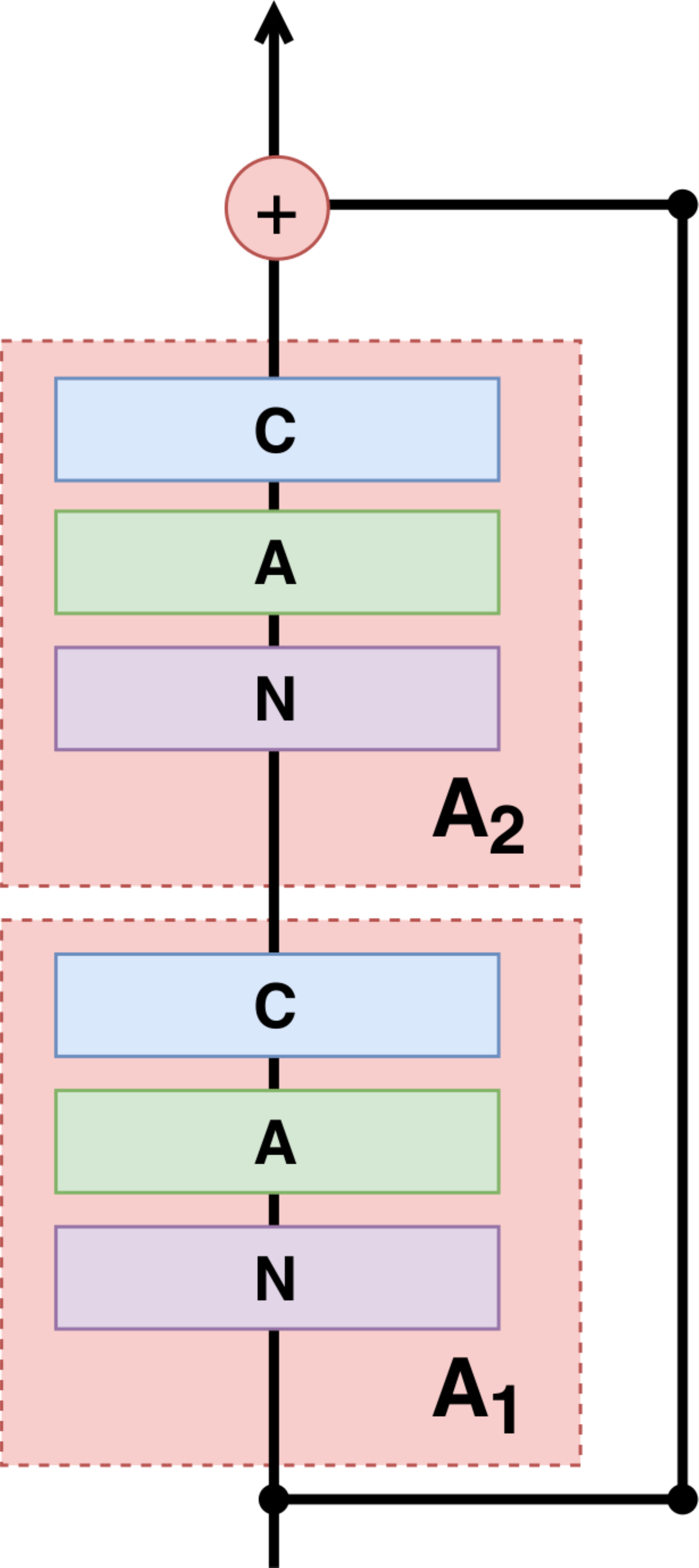}}
\caption{
(a) ResNet  with expert adapters before all blocks. A layer of experts is placed before every block.
(b) Each individual adapter including the overall skip connection.
\mbox{N, A, C} stand for (Group) Normalization, (ReLU) Activation,
and Convolution layers, respectively.
\label{fig:architecture}}
\vspace{-2em}
\end{figure}

\Cref{fig:resnet_with_adapters} depicts the ResNet architecture with multiple expert
adapters ($a_1^{(i)}, \ldots, a_n^{(i)}$). Let $F_i$ be the function implemented by 
the $i$-th block of the backbone network. We \emph{adapt its input} by computing the output as $x_i \coloneqq F_i(x_{i-1} + a_e^{(i)}(x_{i-1}))$,
where $e = R(x)$ is the identity of the selected expert, given by some routing function $R$, 
and $x$ is the original input. During upstream training, the function $R$ may also 
use the labels in addition to the image, as we discuss in  \cref{subsec:expert_training}.

\Cref{fig:adapter_arch} shows the adapter's bottleneck architecture.
An adapter sequentially applies components $A_1$ and $A_2$.
Each component performs a group normalization (N) \cite{Wu_2018_ECCV}, 
a ReLU activation (A) \cite{glorot2011deep}, and a convolution (C) 
\cite{fukushima1980neocognitron,lecun1989backpropagation}, in that order.
Due to the skip connection, the output dimension of $A_2 \circ A_1$ must match that of its input, 
$c$. However, we can change the output channels $k$ of $A_1$, in 
order to limit the amount of parameters. Thus, we set $k = \frac{c}{2}$ so that the 
number of parameters equals that of a linear adapter. Each adapter only increases the parameter count of the R50 backbone by 6\%.
We briefly explored placing these adapters in other locations, or using other variations \cite{rebuffi2018efficient}, but we did not observe any significant improvement.

\subsection{Upstream Data and Expert Definition}

We train our upstream models on large vision datasets
with thousands of classes.
Moreover, the datasets include an expressive hierarchy, linking classes and ancestor concepts via ``is-a'' relationships.
Our experts' domains are nodes in this hierarchy, which are selected automatically based on the number of images.
Due to the multi-label nature of the datasets, several experts could
simultaneously apply to an image.
For example, for an image of a lion, all of
\textit{organism, animal, carnivore, felidae}, and \textit{lion} could be 
relevant expert domains.
In particular, we use two different upstream image datasets, 
and independently train a set of experts on each.
We further describe them in \cref{subsec:exp_upstream}.

\subsection{Expert Training}
\label{subsec:expert_training}
Recall we denote by $\mathbf{B}$ the baseline R50 model trained on the whole upstream dataset $\mathbf{D}_U$.
As shown in \cref{fig:algorithm_overview}, the second step of upstream training consists of training each expert individually on 
different subsets of the upstream dataset.
Let $\mathbf{D}_e := (X_e, Y_e) \subseteq \mathbf{D}_U$ be the data corresponding to expert $e$.
The subsets corresponding to different experts may overlap (e.g.\ for the \emph{animal} and \emph{dog} experts).

As mentioned before, the \emph{full} experts completely fine-tune $\mathbf{B}$ on $\mathbf{D}_e$.
For the \emph{adapter} experts the weights corresponding to the 
adapter $e$ (modules in red in \cref{fig:architecture}) are trained on $\mathbf{D}_e$, but the shared blocks and head parameters are 
\emph{frozen}.
Note that, due to the sharing scheme, we can train all experts independently in parallel.
We train all experts for the same number of steps, regardless of the size of $\mathbf{D}_e$.
Instead of learning a routing function, we exploit the structure of the upstream labels and
use a hard-coded routing. 
We found this makes learning easier, and leads to powerful specialized models.

\section{Expert Selection}
\label{sec:transfer}

Given a new downstream dataset $\mathbf{D}_T = (X_T, Y_T)$, we must choose an expert to use.
We consider three approaches: \emph{domain prediction}, \emph{label matching}, and \emph{performance proxy}.

\textbf{Domain Prediction.}
This strategy selects the expert solely based on the images $X_T$. It effectively selects the expert whose domain best matches the target images.
We implement this by training an auxiliary network (the ``Expert Prediction Network'' or EPN) to classify the expert from the image (i.e.\ learn the hard-coded routing mentioned previously).
The EPN is trained upstream using the pre-training data and expert assignments.
During transfer, an expert is selected using the highest geometric mean EPN predictive probability across the dataset.
Details are in the \Cref{app:epn}.%

\textbf{Label Matching.}
Alternatively, matching of the expert to the task can be done in the label space as opposed to the input space.
This approach is similar in spirit to the one described in~\cite{ngiam2018domain}.
We implement this strategy by computing the affinity of each expert to a new downstream task in the label space of the upstream dataset.
We first use a generic network trained on all upstream labels to predict upstream labels on the downstream images.
We compute the KL-divergence between the distribution of labels on the downstream task images, and the prior distribution of labels for each expert.
This per-expert prior is computed as the empirical distribution of labels on the images used to train that expert.
We select the expert with the smallest KL-divergence.
Details are in the \Cref{app:kl_routing}.

\textbf{Performance Proxy.}
The aforementioned two strategies are simple, but do not use the training labels $Y_T$ available for downstream tasks, which may contain key information.
It would be too expensive to fine-tune every expert to every new task and select the best with hindsight, so we propose a proxy for the final performance.
For this, we use a $k$-nearest neighbors classifier \cite{altman1992introduction} with the %
image embeddings produced by each expert.
In the case of full experts, we simply apply the corresponding full network to compute these embeddings.
For adapter-based experts, we apply the specific expert and ignore the remaining ones.
Concretely, let $M_e(x)$ be the embedding corresponding to expert $e$ on input $x$, and let $\mathbf{D}_T = \{ (x_i, y_i)_{i=1}^{N_T} \}$ be our downstream task.
In order to score each expert, we apply a kNN classifier on the embedded dataset $\mathbf{D}_{T,e} = \{ (M_e(x_i), y_i)_{i=1}^{N_T} \}$, with $k=1$ and Euclidean distance.
The accuracy $\mathrm{acc}(\mathbf{D}_{T,e})$ is computed via leave-one-out cross-validation.
Finally, we select the expert with highest accuracy: $e^* = \arg\max_e \mathrm{acc}(\mathbf{D}_{T,e})$.
There are other alternative proxies that are cheaper than full fine-tuning, for example fitting a logistic regression, SVM, or decision trees to the features.
These proxies may better match final performance.
However, we elect to use a kNN since it is computationally cheap ---
it only requires a forward pass through the data, and leave-one-out cross-validation requires no additional inference per-fold ---  and it performs well (\cref{sec:experiments}).

\subsection{Downstream transfer}
The expert selection algorithm could choose several experts to
be combined to solve any target task. 
However, we limit the scope of our work to transferring a single expert per task, since this approach is simple and turns out to be effective.
Thus, the downstream transfer procedure is straightforward: it simply involves fine-tuning 
the selected expert model.
We fine-tune the entire expert network to the downstream dataset, including the adapters when applicable.
This differs from the original residual adapters work~\cite{rebuffi2017learning}, where only the adapters were fine-tuned -- 
when we tried this, it performed poorly.
While it was valuable to restrict the scope of upstream training to focus on specializing the expert adapter parameters, we found fine-tuning the whole network downstream to be greatly beneficial.

\section{Experimental Results}
\label{sec:experiments}

\subsection{Upstream Training}
\label{subsec:exp_upstream}
We train experts using two large datasets with hierarchical label spaces.

\textbf{ImageNet21k} \cite{deng2009imagenet} is a public dataset containing 13 million images, and 14 million labels of 21\,843 classes, which are WordNet synsets \cite{fellbaum2012wordnet}.
In addition to the 21k classes, we consider the 1\,741 synsets that are their ancestors.
We use the 50 synsets of ImageNet21k with the largest number of images to train the expert models. 
These include e.g.\ \emph{animal, artifact, organism, food, structure, person, vehicle, plan}, or \emph{instrument}.

\textbf{\JFT} is an even larger dataset of 300M images used in~\cite{chollet2017xception, hinton2015distilling, ngiam2018domain, sun2017revisiting},
containing 300 million images and 18\,291 classes.
Each image can belong to multiple classes, and as for ImageNet21k, the classes are organized
in a hierarchy.
We select as expert domains the classes with a sufficiently large number of examples:
the 240 classes with more than 850\,000 images. Some of the automatically selected experts are 
\emph{animal, arts, bird, food, material, person, phenomenon, plant}, or \emph{product}.

We pre-train generic models on a Cloud TPUv3-512, using the same protocol as 
\cite{kolesnikov2019large}. Then fine-tune them briefly on each slice to create the expert models.
Additional details are found in \cref{sec:upstream_training_details}.

\subsection{Downstream Tasks}

We evaluate on two suites of tasks, each consisting of several datasets.
The first is the Visual Task Adaptation Benchmark (VTAB) \cite{zhai2019visual}, which consists of 19 datasets.
We evaluate on VTAB-1k, where each task contains only 1k training examples.
The tasks are diverse, and divided into three groups: natural images (single object classification), structured tasks (count, estimate distance, etc.), and specialized ones (medical, satellite images).
\Cref{sec:vtab_data_details} contains further details.

The second suite is a collection of popular natural datasets commonly used in transfer learning literature:
FGVC-Aircraft~\cite{maji13fine-grained}, 
Birdsnap~\cite{berg-birdsnap-cvpr2014},
CIFAR10~\cite{Krizhevsky09learningmultiple}, 
Stanford Cars~\cite{KrauseStarkDengFei-Fei_3DRR2013}, 
Food~\cite{bossard14}, and 
Oxford IIIT Pets~\cite{parkhi12a}.
Oxford IIIT Pets is also part of the Visual Task Adaptation Benchmark.

\subsection{Transfer Evaluation Protocol}
\label{subsec:eval_protocol}
When transferring to new tasks we need to perform expert selection and choose other hyperparameters (e.g. learning rate for fine-tuning).
For each downstream task, we use the following three step protocol.

\textbf{Expert Transfer}. 
We select the expert to transfer using one of the methods presented in \cref{sec:transfer}.
In both sets of tasks, we use 1k training examples per dataset.
Details are provided in \cref{subsec:knn_hyperparams}.

\textbf{Hyperparameter Selection}.
In VTAB-1k we use the recommended hyperparameter sweep and 800-training/200-validation split in \cite{zhai2019visual}. We independently repeat the hyperparameter selection procedure 10 
times for confidence intervals. %
For the other datasets we perform a single random search over 36 hyperparameter sets and select
the best set based on the validation performance.
This is a similar computational budget to that of \cite{ngiam2018domain}.
See \cref{subsec:vtab_hyperparams,subsec:domain_adaptive_hyperparms} for sweep details.

\textbf{Final Re-training}.
Using the hyperparameters from the previous step, we re-train the selected expert on the entire task 
(training plus validation set). 
In VTAB-1k, we repeat this step 3 times for each 
of the 10 trials of hyperparameter selection and compute the test
accuracy, yielding 30 outcomes per method per task.
We compute the {median} of these 30 outcomes as the final accuracy in the dataset.

\subsection{Performance of Different Expert Selection Strategies}
\label{subsec:experiments_selection_algorithm}

We first establish which of the expert selection strategies presented
in \cref{sec:transfer} performs best. As a baseline we also try 
selecting a random, uniformly drawn, expert per task.
\Cref{tab:vtab_test_transfer_algorithms} shows the results on VTAB-1k,
using full experts trained on \JFT.
\Cref{tab:selection_algorithms_full_and_adapters} show the results with adapters.

\begin{table}[tb]
\centering
\caption{VTAB-1k results of different selection algorithms, using full experts
trained on \JFT. The average accuracy across each group of tasks and
across all VTAB is reported. In each dataset, the median accuracy over 30 runs 
is used. Bootstrapped confidence intervals at 95\% level are included.
\label{tab:vtab_test_transfer_algorithms}}
\begin{tabular}{lllll}
\toprule
& \textsc{Natural} & \textsc{Specialized} & \textsc{Structured} & \textsc{All}\\
\midrule
Random
& \cih{60.6}{59.1}{63.9} 
& \cih{81.2}{80.9}{81.8} 
& \cih{56.8}{54.9}{57.8} 
& \cih{63.3}{62.3}{64.6} \\
Domain Prediction
& \cih{75.9}{74.4}{77.4} 
& \cih{81.5}{81.3}{82.2} 
& \cih{\textbf{57.0}}{56.1}{57.4} 
& \cih{69.1}{68.4}{69.8} \\
Label Matching
& \cih{77.6}{77.8}{78.1} 
& \cih{80.3}{79.1}{82.5}
& \cih{56.9}{55.6}{57.2} 
& \cih{69.6}{68.9}{70.0} \\
Performance Proxy
& \cih{\textbf{79.7}}{79.5}{80.0} 
& \cih{\textbf{83.6}}{83.3}{83.8} 
& \cih{55.3}{52.1}{56.3} 
& \cih{\textbf{70.2}}{68.9}{70.6}\\
\bottomrule     
\end{tabular}
\end{table}

Overall, all methods perform better than random selection, particularly on the 
\textsc{Natural} group. This confirms that selecting good experts is essential. Overall, the performance proxy (kNN) selection performs better than the other alternatives.
kNN's average accuracy is 11\% (relative) and 5.5\% higher than that of the domain prediction 
and label matching, respectively. 
Thus, making use of the downstream labels 
offers a significant advantage in expert prediction.
Therefore, in all subsequent experiments we use the kNN-based selection. 
We did not see a strong difference for the \textsc{Structured} datasets.
We provide an extensive analysis of the kNN accuracy distribution per expert in \cref{app:knn}.
\Cref{app:random_experts} shows how training experts on \emph{random} subsets of the upstream data does not work well.

\vspace{-0.8em}
\subsection{Results on VTAB}
\label{subsec:experiments_architectures}

\begin{table}[tb]
\centering
\caption{VTAB-1k results of the baseline models and different expert architectures using kNN selection,
pre-trained on ImageNet21k (IN21k) and \JFT. The average accuracy across each group of tasks and 
across all 19 tasks is shown. In each dataset, the median accuracy over 30 runs is used.
\label{tab:vtab_all_results}}
\begin{tabular}{llllll}
\toprule
    & & \textsc{Natural} & \textsc{Specialized} & \textsc{Structured} & \textsc{All} \\
\midrule
\multirow{4}{*}{IN21k}
& Baseline 
& \cih{77.7}{77.4}{77.8} 
& \cih{82.0}{78.4}{83.9} 
& \cih{56.8}{55.9}{57.2}
& \cih{69.8}{68.8}{70.3} \\
& Adapters 
& \cih{78.1}{78.0}{78.3} 
& \cih{83.5}{83.1}{83.6}
& \cih{57.5}{56.8}{58.2}
& \cih{70.6}{70.3}{70.9} \\
& Full 
& \cih{78.3}{78.1}{78.6} 
& \cih{83.4}{83.2}{83.6} 
& \cih{59.4}{58.7}{59.8} 
& \cih{71.4}{71.1}{71.6} \\
& All Experts 
& \cih{78.3}{78.1}{78.6} 
& \cih{83.6}{83.4}{83.7} 
& \cih{58.8}{58.0}{59.4} 
& \cih{71.2}{70.8}{71.5} \\
\midrule
\multirow{4}{*}{JFT}
& Baseline 
& \cih{77.4}{77.3}{77.6} 
& \cih{81.6}{81.5}{82.0} 
& \cih{57.2}{52.8}{58.2} 
& \cih{69.8}{68.0}{70.2} \\
& Adapters 
& \cih{79.0}{78.6}{79.1} 
& \cih{81.3}{79.2}{82.5} 
& \cih{59.1}{58.3}{60.1} 
& \cih{71.1}{70.5}{71.6} \\
& Full 
& \cih{79.7}{79.5}{80.0} 
& \cih{83.6}{83.3}{83.8} 
& \cih{55.3}{52.2}{56.2} 
& \cih{70.2}{68.9}{70.6} \\
& All Experts
& \cih{80.0}{79.2}{80.4} 
& \cih{83.7}{83.6}{83.8} 
& \cih{58.6}{58.0}{59.4} 
& \cih{71.8}{71.3}{72.2} \\
\midrule
IN21k + JFT
& All Experts 
& \cih{\textbf{80.2}}{79.8}{80.3} 
& \cih{\textbf{84.0}}{83.7}{84.2} 
& \cih{\textbf{59.5}}{58.7}{60.1} 
& \cih{\textbf{72.3}}{71.9}{72.6} \\
\bottomrule  
\end{tabular}
\vspace{-1em}
\end{table}    

\Cref{tab:vtab_all_results} shows the average accuracy across all the 19 VTAB-1k datasets broken down 
by type (natural, specialized, and structured). 
We summarize our findings as follows:

\textbf{Improvement over Non-expert Baseline.}
All the algorithms, trained on either \JFT or ImageNet21k, improve their corresponding Baseline on VTAB.
he results are most pronounced on the \textsc{Natural} datasets.
While we also see improvements in \textsc{Specialized} and \textsc{Structured} datasets, some of the confidence intervals overlap.
The performance of both \JFT and ImageNet21k models is fairly similar in general.
This is not unexpected; it has been observed before that, with restricted model capacity, \JFT and ImageNet21k perform very similarly~\cite{kolesnikov2019large}.
\Cref{app:selected_experts} shows the selected experts.

\textbf{Quality of Natural Representations.}
The upstream datasets used to train the experts mostly contain natural images.
Consequently, the spectrum of representations offered by our models seem very effective in downstream natural datasets.
More concretely, all models lead to improvements over the baseline performance, with average gains ranging from 1\% to over 3.3\% on the 7 natural datasets.

\textbf{Full vs.\ Adapters.}
\emph{JFT Experts.}
Full models outperform adapters convincingly in \textsc{Natural} and \textsc{Specialized} datasets.
However, they do a poor job on \textsc{Structured} datasets --mainly due to the failure on one specific dataset.
 \emph{ImageNet21k Experts.}
In this case, the advantage of full experts comes precisely from \textsc{Structured} datasets.
\Cref{sec:vtab_details} provides results broken down by each dataset.

\textbf{Combining All Experts.} The previous numbers suggested combining all experts (full or adapter trained on JFT or ImageNet -- almost 600 models). The results are remarkable: the mean relative improvement over the Baseline across all 
VTAB datasets is 3.6\%, %
showing gains on all dataset types.

\subsection{Our Approach vs. Domain Adaptive Transfer}
\label{subsec:adaptive_transfer_comparison}

Domain Adaptive Transfer~\cite{ngiam2018domain} (DAT) also relies on specialist models pre-trained on JFT.
First it trains a generalist model on the upstream data, 
similar to our $\mathbf{B}$.
For any new task, then re-weights the \emph{upstream} images based on 
a forward pass on the downstream data, and fine-tunes a new specialist
model using the re-weighted upstream data. Finally, 
the model is further tuned on the target task.
DAT falls outside of our transfer setup presented in 
\cref{sec:transfer},
as the downstream data \emph{directly} influences the upstream training.
This incurs a significant upstream cost to learn \emph{every} new
target task.

Remarkably, our algorithm works in setups where access to upstream data is not available (e.g. for privacy or proprietary reasons).
We also use downstream labels, which proved to carry key information about the task (see \cref{subsec:experiments_selection_algorithm}). 
And most importantly, our method is more practical by amortizing the cost of expert pre-training as more downstream tasks are served. 
Under same models and hardware, running kNN (with 240 models) is between $500\times$--$1000\times$ faster than fine-tuning the 
baseline model with the re-weighted upstream data. \Cref{sec:domain_adaptive_details} has additional details. 

\begin{table}[tb]
\centering
\caption{Accuracy on the datasets in~\cite{ngiam2018domain},
and the average accuracy
across the six of them. Bootstrapped confidence intervals at 95\% level are 
shown next to the accuracy where available.
\cite{ngiam2018domain} report results using Inception-v3 (In-v3) and a larger network, AmoebaNet-B (Am-B).}
\label{tab:dat_results}
\setlength{\tabcolsep}{2pt}
\resizebox{1.0\textwidth}{!}{%
\begin{tabular}{llllllll}
\toprule
& \textsc{Aircraft} & \textsc{Birds} & \textsc{Cars} & \textsc{CIFAR10} & \textsc{Food} & \textsc{Pets}* & \textsc{Avg.}\\    
\midrule  
Baseline
  & \cih{91.4}{91.0}{91.7} & \cih{78.8}{78.0}{79.4} & \cih{95.6}{95.4}{95.7} & \cih{97.8}{97.7}{97.9} & \cih{91.3}{91.2}{91.5} & \cih{94.5}{94.4}{94.6} & \cih{91.6}{91.4}{91.7} \\
Adapters (JFT)
  & \cih{92.5}{92.2}{92.8} & \cih{79.4}{78.7}{80.1} & \cih{95.9}{95.8}{96.0} & \cih{97.9}{97.8}{98.0} & \cih{91.6}{91.5}{91.7} & \cih{94.6}{94.4}{94.8} & \cih{92.0}{91.9}{92.1} \\
Full (JFT)
  & \cih{\textbf{94.8}}{94.5}{95.1} & \cih{{83.6}}{83.1}{83.9} & 
  \cih{\textbf{96.1}}{96.0}{96.3} & \cih{97.8}{97.7}{97.9} & 
  \cih{93.1}{92.8}{93.2} & \cih{\textbf{97.0}}{96.9}{97.1} & 
  \cih{93.7}{93.6}{93.8} \\
\midrule
Dom-Ad (In-v3) \cite{ngiam2018domain}
  & 94.1 & 81.7 & 95.7 & {98.3} & {94.1} & \textbf{97.1} & 93.5 \\ 
Dom-Ad (Am-B) \cite{ngiam2018domain}
  & 92.8 & \textbf{85.1} & 95.8 & \textbf{98.6} & \textbf{95.3} & {96.8} & \textbf{94.1} \\ 
\bottomrule
\end{tabular}
}
\begin{flushleft}
\small{*Pets results are mean per class accuracy as opposed to mean accuracy.}\end{flushleft}\vspace{-0.5cm}
\end{table}

\Cref{tab:dat_results} shows the mean accuracy over 30 trials per dataset, on the same datasets and under a similar hyperparameter budget as DAT.
These tasks are close to VTAB's \textsc{Natural} group 
and yield similar results: full experts outperform adapters.
A number of differences make our results not directly comparable to DAT.
In particular, they use Inception-v3~\cite{szegedy2016rethinking}, and AmoebaNet-B~\cite{real2019regularized} models.
Inception-v3 and R50 are similar in performance and size; the former has 24M parameters, attaining 78.8\% top-1 on ILSVRC2012 (from-scratch), whereas the latter has 26M parameters and attains 76.0\%.
The AmoebaNet-B (N=18, F=512) is 22 times larger, with more than 550M parameters.
Despite the differences, our method is competitive and matches or beats DAT in half the datasets.

\vspace{-0.5em}
\section{Discussion}
\textbf{Algorithm.}
Our results suggest that there are strong potential benefits to using smartly routed pre-trained experts \emph{when} the domain of the experts broadly matches that of the downstream tasks.
We have clearly seen this with natural images.
Instead, as expected, when there is a skill mismatch (e.g.\ trying to solve a counting task with diverse single-object recognition experts) we have not observed any significant gain or loss.
Still, in these cases, the expert selector can fall back on the generic model or representation.
When there is an extremely relevant expert for a task --say, our \textit{flower} or \textit{plant} models for the Oxford Flowers 102 task--, using full network experts proved beneficial.
In contrast, many datasets did not have a perfect match, and adapters seemed easier to fine-tune in these cases. 

\textbf{Impact.}
In the near future, we foresee large computer vision systems composed by a wide range of pre-trained specialist modules.
These modules may be based on huge amounts of data, small but high-quality curated repositories, or even on private and proprietary content, and they would cover a diverse spectrum of canonical tasks (object recognition, some way of narrow reasoning, counting, sorting, etc.).
Some of them may not even need to be end-to-end learned from data.

\textbf{Future Directions.}
There are a number of exciting follow-up research directions.
Selecting and combining multiple experts for any downstream task is a natural extension of our work.
This could be especially useful for tasks that require understanding several concepts, not necessarily captured by a single expert.
Per-example routing (i.e. applying routes tailored to each individual data-point) could also lead to improvements based on targeted processing, for example, in the context of tasks with instances of various difficulties.
Finally, moving beyond our experts based on label hierarchies, and towards automatic discovering and training of experts could unlock even further gains.

\begin{ack}
We would like to thank Josip Djolonga and Wenlei Zhou for useful comments, feedback, and remarks.
\end{ack}

\bibliography{main}
\bibliographystyle{abbrv}

\clearpage
\appendix

\section{Expert Predictor Networks}
\label{app:epn}
An expert predictor network (EPN) tries to directly predict the relevant expert
for an input image, using only the image itself as input.
We first train the EPN upstream, and then we apply it downstream to select the most relevant expert for a new task by aggregating its output on all the images in the task.

\textbf{EPN Upstream Training.} As we described in \cref{sec:upstream}, we have split the upstream dataset $D_U$ into a
collection of subsets $\{D_e : 1 \leq e \leq E\}$, with $D_e = (X_e, Y_e) \subseteq D_U$. 
In order to train the EPN, we simply assign the expert identity $e$ as the label of all
images in $X_e$, and train the network in a supervised manner (using softmax cross-entropy) to predict the expert.
Expert slices $\{D_e\}$ are not disjoint, thus, it is
possible that an individual image appears multiple times in the training data for the EPN 
with different expert identities. Because subsets sizes are different, and in order not to favor any particular expert, we resample the training images so that each expert is seen equally often.
Intuitively, this classification problem should be substantially easier than predicting
the upstream classes $y$ directly, as there are much fewer experts than upstream classes.

\textbf{Downstream Expert Selection.} Suppose we are given a downstream task, containing images $X_T = \{\mathbf{x}_1, \ldots, \mathbf{x}_{N_T}\}$.
We first apply a forward pass on $X_T$ using the EPN.
Let $Q_{\EPN}(e \mid X = \mathbf{x}_i)$ be the probability assigned by the EPN to expert $e$ for input image $\mathbf{x}_i$.
In order to make a single decision for the whole downstream task, we combine those probabilities using a log-linear transformation.

We select the expert as follows:
\begin{equation}
\hat{e} = 
\argmax_{e} \frac{1}{N_T} \sum_{i=1}^{N_T} \log Q_{\EPN}(e \mid X = \mathbf{x}_i).
\end{equation}

The log-linear combination of per-example probabilities was obtained after several experiments with a number of functions.
Intuitively, this transformation penalizes experts that only apply to a 
subset of the downstream data, but are not relevant to other downstream examples.

A major drawback of the EPN is the fact that it does \emph{not} use or benefit from the 
downstream labels.
Imagine there is an image dataset with pictures containing simultaneously both lions and elephants.
Suppose we are faced with two different downstream tasks based on the same inputs: one is to count lions, the other is to count elephants.
Furthermore, imagine our experts happen to include \emph{lion} and \emph{elephant}.
Depending on the task, it would be reasonable to choose one or the other expert.
Unfortunately, the basic EPN approach is agnostic to the outputs, and --as the input images are identical-- it would return the same selected expert in both cases.

\section{Kullback--Leibler divergence}
\label{app:kl_routing}

Since our expert datasets $\mathbf{D}_e$ were built based on the hierarchy of labels in the
upstream dataset, it is reasonable to assume that the prior distribution of the
labels in each $\mathbf{D}_e$ differ across experts $e$. Let $P_e$ be this prior 
distribution for expert $e$. Then, we can use a divergence measure, such as the Kullback--Leibler (KL), to determine which expert to use. If one assumes that the 
downstream dataset is well represented by the upstream dataset $\mathbf{D}_U$ (although not 
necessarily by an individual $\mathbf{D}_e$), one can use the baseline neural network $\mathbf{B}$ to 
approximate the distribution of \emph{upstream} labels conditioned to the set of 
downstream images:
\begin{equation}
Q(Y) \coloneqq \frac{1}{N_T} \sum_{j=1}^{N_T} Q_{\mathbf{B}}(Y \mid X = \mathbf{x}_j),
\end{equation}
where $Q_{\mathbf{B}}(Y \mid X = \mathbf{x}_j)$ is the probability distribution given by
$\mathbf{B}$ over each image in the set $X_T = \{\mathbf{x}_1, \ldots, \mathbf{x}_{N_T}\}$ 
of downstream images.
Then, we simply select the expert with the lowest KL divergence:
\begin{equation}\label{eq:kl_divergence_selection}
\hat{e} \coloneqq \argmin_{e} \divkl{P_e}{Q}
\end{equation}

This allows us to leverage the baseline model that we already trained, and 
not train an auxiliary neural network to predict the expert to use, like the
EPN in \cref{app:epn} does. In addition, this has the benefit of using information about the distribution of upstream classes, which may be useful when the target classes
are well represented among the upstream ones.

In our case, the upstream datasets consists of multi-labeled images. 
The distribution of labels given to a particular image is modelled by the neural
network as a joint distribution of independent Bernoulli random variables.
Assuming this independence also holds for $P_e$, one can then compute 
\cref{eq:kl_divergence_selection} very efficiently.
Of course, this assumption is not true in either case
(e.g. the presence/absence of the \emph{dog} and \emph{animal} labels is not 
independent), but it is standard practice for multi-label classification.
\section{Further Results on \texorpdfstring{$k$}{k}-Nearest Neighbors}
\label{app:knn}

In this section, we present the kNN accuracy distribution per dataset that we found for both \JFT and ImageNet21k experts.
A flat curve indicates differences across experts may not be very relevant for the downstream task, while steep regions suggest strong decreases in value among expert models.

\subsection{kNN Hyperparameters}
\label{subsec:knn_hyperparams}

We select the expert to transfer using the kNN transfer proxy with $k=1$ and a Euclidean distance metric. We use 1\,000 training examples 
in all datasets (including those in the comparison with Domain 
Adaptive Transfer \cite{ngiam2018domain}), and compute kNN 
leave-one-out cross-validation to compute the accuracy per expert.
Finally, we select the one with highest accuracy.
For VTAB-1k this corresponds to the entire training set per task; 
whereas for the other tasks, we randomly sample 1k training examples.
We do not perform special data pre-processing, and simply resize and crop to $224 \times 224$, as done in upstream evaluation.

We used a NVDIA V100 GPU to perform the kNN selection for each dataset, with this hardware, 
selecting among 240 models takes less than 2 hours.

\subsection{Architecture Comparisons}
\label{app:arch}

\begin{figure}[htb]
\includegraphics[width=\textwidth]{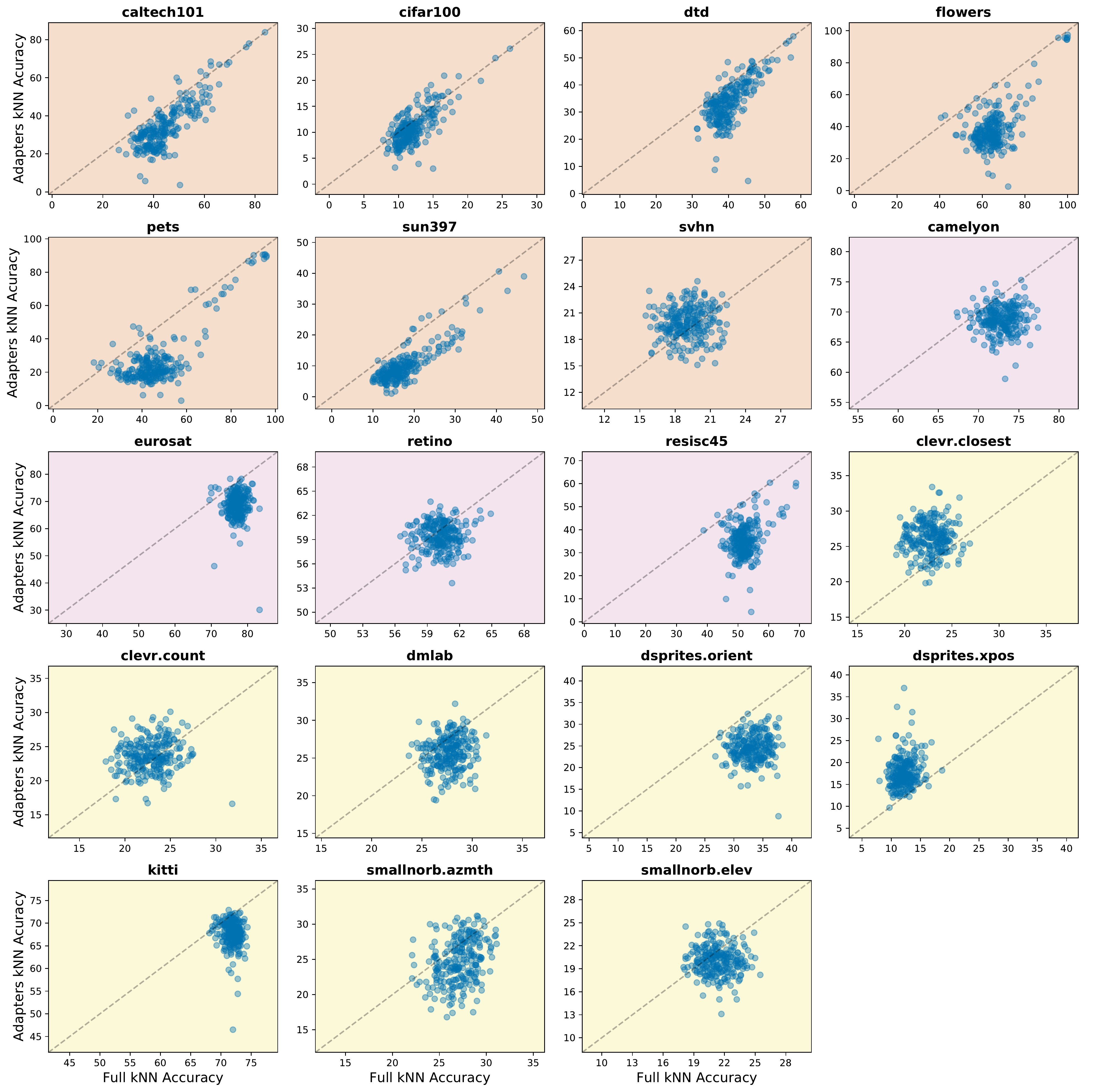}
\caption{\textbf{JFT Experts}.
Each point is one expert (upstream data slice); the $x$-axis represents the kNN accuracy before downstream finetuning of the expert trained on a \emph{full} network. The $y$-axis displays the kNN accuracy before downstream finetuning of the expert trained with an \emph{adapter} module.
The background color indicates the dataset group: 
{\naturalsym~\textsc{Natural}}, 
{\specializedsym~\textsc{Specialized}}, and
{\structuredsym~\textsc{Structured}}.
There are 244 experts.
Identity dashed line shown too.}
\label{fig:app_jft_per_expert_knn}
\end{figure}

\begin{figure}[htb]
\includegraphics[width=\textwidth]{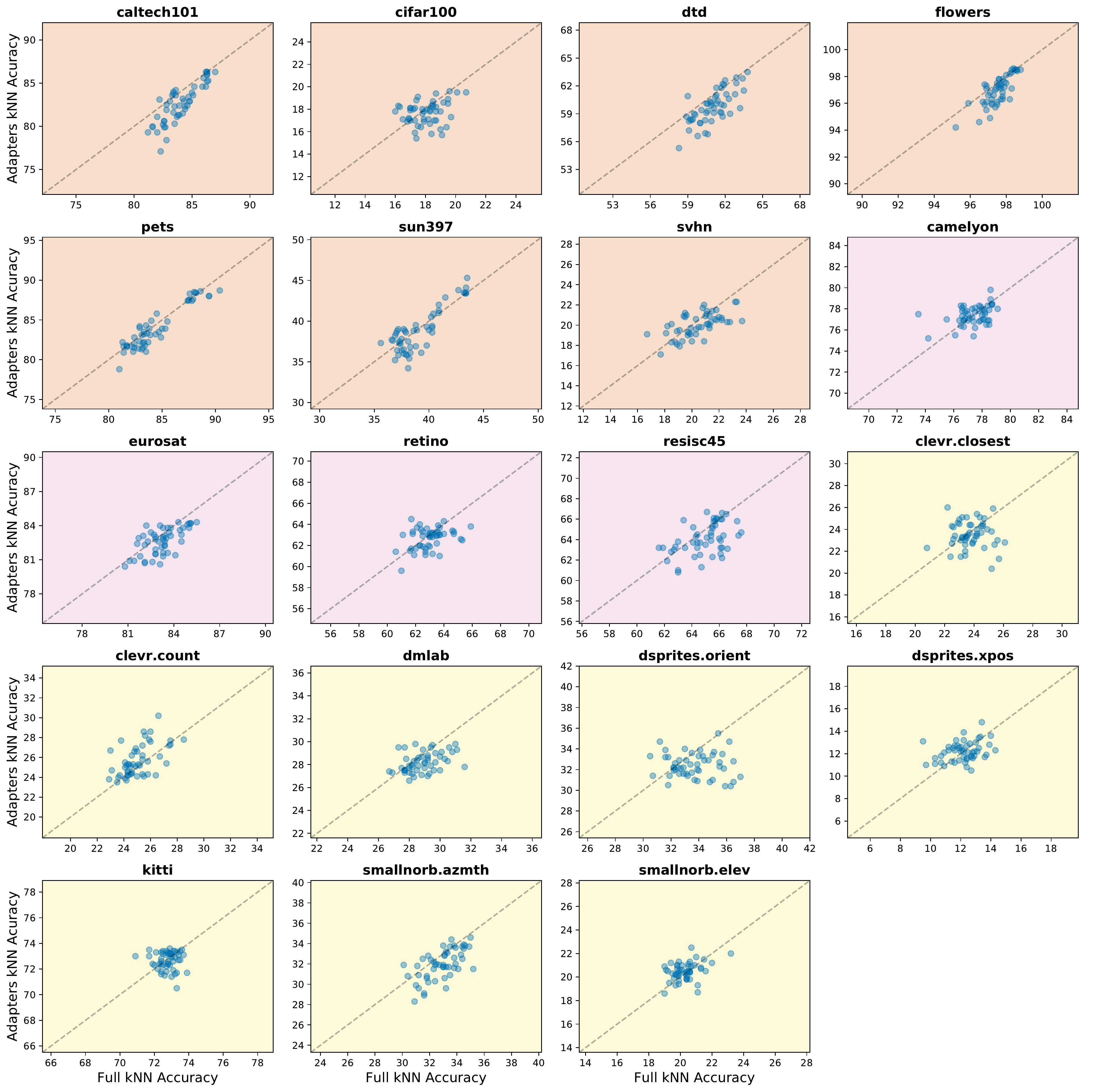}
\caption{\textbf{ImageNet21k Experts}.
Each point is one expert (upstream data slice); the $x$-axis represents the kNN accuracy before downstream finetuning of the expert trained on a \emph{full} network. The $y$-axis displays the kNN accuracy before downstream finetuning of the expert trained with an \emph{adapter} module.
The background color indicates the dataset group: 
{\naturalsym~\textsc{Natural}}, 
{\specializedsym~\textsc{Specialized}}, and
{\structuredsym~\textsc{Structured}}.
There are 50 experts. Identity dashed line shown too.}
\label{fig:app_inet_per_expert_knn}
\end{figure}

In this section we look at the kNN accuracy \emph{before} transferring the experts, and --in particular-- at how it depends on the architecture choice.
Recall each expert is associated with one slice of the upstream data.
For any given slice, we have trained both a full ResNet50 network, and adapters attached to a pretrained ResNet50.
We plot the accuracy achieved by these representations (scatter-plot, full at $x$-axis; adapters at $y$-axis) for all experts for each group of VTAB datasets.
We look at JFT experts (\Cref{fig:app_jft_per_expert_knn}) and at ImageNet21k experts (\Cref{fig:app_inet_per_expert_knn}).
Ideally, we would expect some positive correlation if expert representations were somewhat similar regardless of the architecture.

\textbf{JFT Experts.}
The first seven plots in \Cref{fig:app_jft_per_expert_knn} show the results in natural datasets.
While most experts do not seem relevant --and performance seems a bit uncorrelated between both types of models--, in most datasets we see that there are a few good experts (top right corner) which offer the strongest performance despite of the selected architecture.
SVHN seems to be an exception.

Similar plots are displayed in the following 4 and the last 8 plots in \Cref{fig:app_jft_per_expert_knn} for specialized and structured datasets, respectively.
Few datasets show agreement on the most promising expert slices, such as Eurosat or Resisc45.
Unfortunately, there is no clear agreement in most specialized and structured datasets.

\textbf{ImageNet21k Experts.} 
We see a reasonable agreement among the best ImageNet21k experts in natural datasets (see first 7 plots in \Cref{fig:app_inet_per_expert_knn}).
While not as correlated as in the case of natural datasets, we still see some positive relationship in some specialized (Eurosat, Resisc45, Patch Camelyon) and structured (Clevr Count, DSprites Position, Smallnorb Azimuth) datasets.

\subsection{kNN Accuracy Distribution for JFT Experts}

\begin{figure}[tb]
\centering
\includegraphics[width=\textwidth]{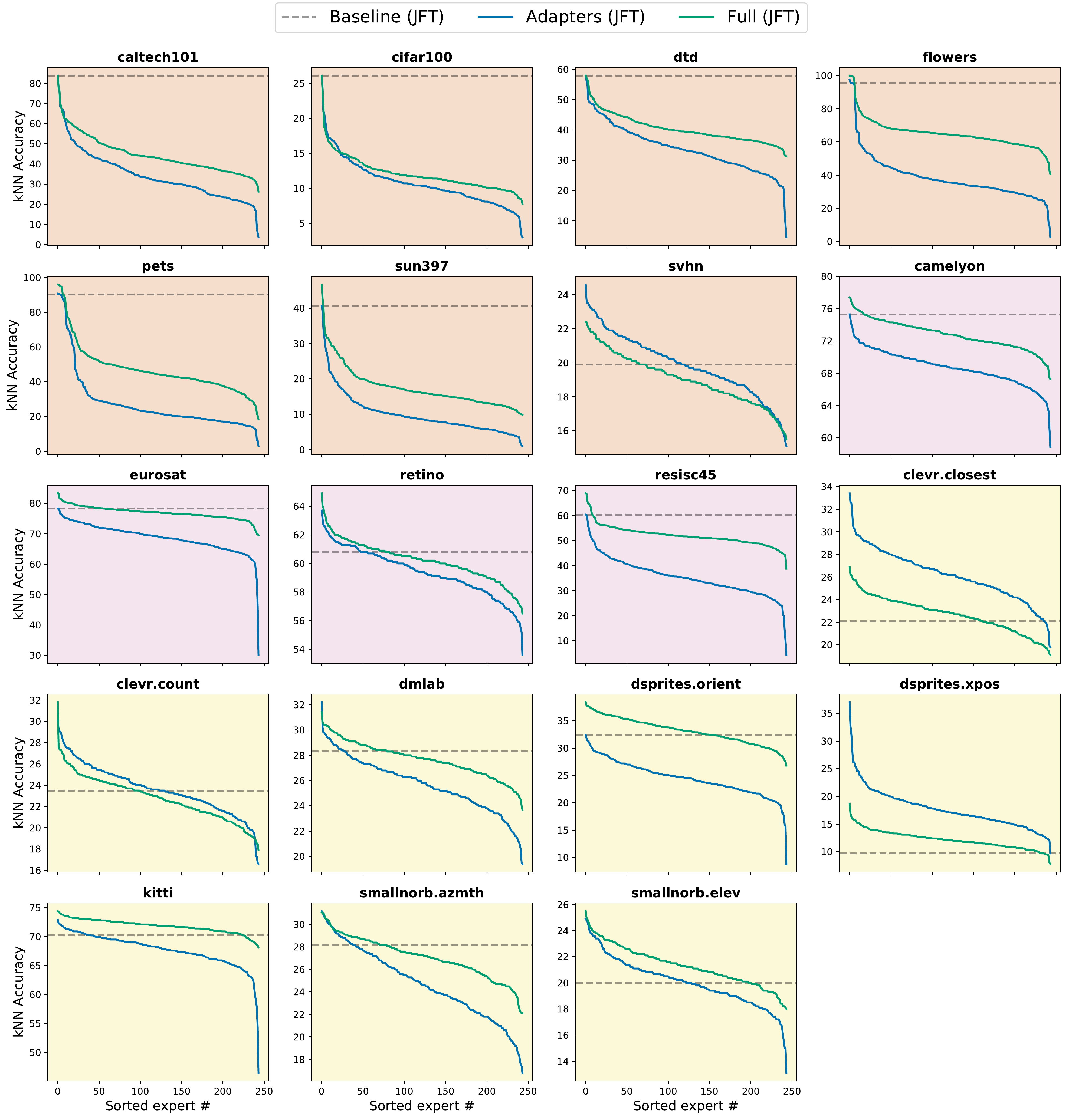}
\caption{Distribution of the kNN accuracy from experts trained on JFT. 
The dashed lines shows the kNN accuracy of the baseline model. 
In each dataset, the experts are sorted according to their accuracy.
The background color of the plot represents the group of the dataset:
\naturalsym~\textsc{Natural}; 
\specializedsym~\textsc{Specialized}; 
\structuredsym~\textsc{Structured}.
\label{fig:knn_accuracy_jft_distribution}}
\end{figure}

\Cref{fig:knn_accuracy_jft_distribution} shows the distribution of the kNN 
accuracy obtained from the embedding of each of the experts trained on \JFT.
In each case (full and adapters), the kNN accuracy of the 244 experts has been sorted in a decreasing manner.
Note that we pick the single expert with highest-score (although other approaches are possible).

In most {\naturalsym~\textsc{Natural}} datasets, we observe that full JFT experts are on average better
than their adapter counter-parts.
In datasets like Caltech101, Cifar100, or DTD, it seems these differences do not affect the top experts, 
while in others (such as Flowers, Pets, and Sun397) the differences still apply to the best expert.
Also, overall, we see that in natural datasets there are usually \emph{strong} differences between good 
and bad experts.
The range of kNN accuracies is pretty large for Caltech101, Flowers, or Pets, where 
some experts seem to already solve the task, while others lead to quite poor accuracies.
The latter may be fixed to some extent by downstream fine-tuning.

In the {\specializedsym~\textsc{Specialized}} group we see a similar pattern in the comparison between full and adapter-based experts.
However, the accuracy range of variations (except, maybe, at the very worst end) is narrower.

The story for {\structuredsym~\textsc{Structured}} datasets with JFT experts is a bit different.
In some datasets, adapters models lead on average to better initial representations (such as Clevr Closest and Clevr Count, or dSprites Position).
As with structured datasets, the difference between the best and the worst experts is shorter.
This may be in part explained by the hardness of the task itself (the average accuracy \emph{after} fine-tuning is definitely lower than in the natural case), but there are some counter-examples to this, like dSprites Position where final accuracies go up to around 90\%.

\clearpage
\subsection{kNN Accuracy Distribution ImageNet21k Experts}

\begin{figure}[tb]
\centering
\includegraphics[width=\textwidth]{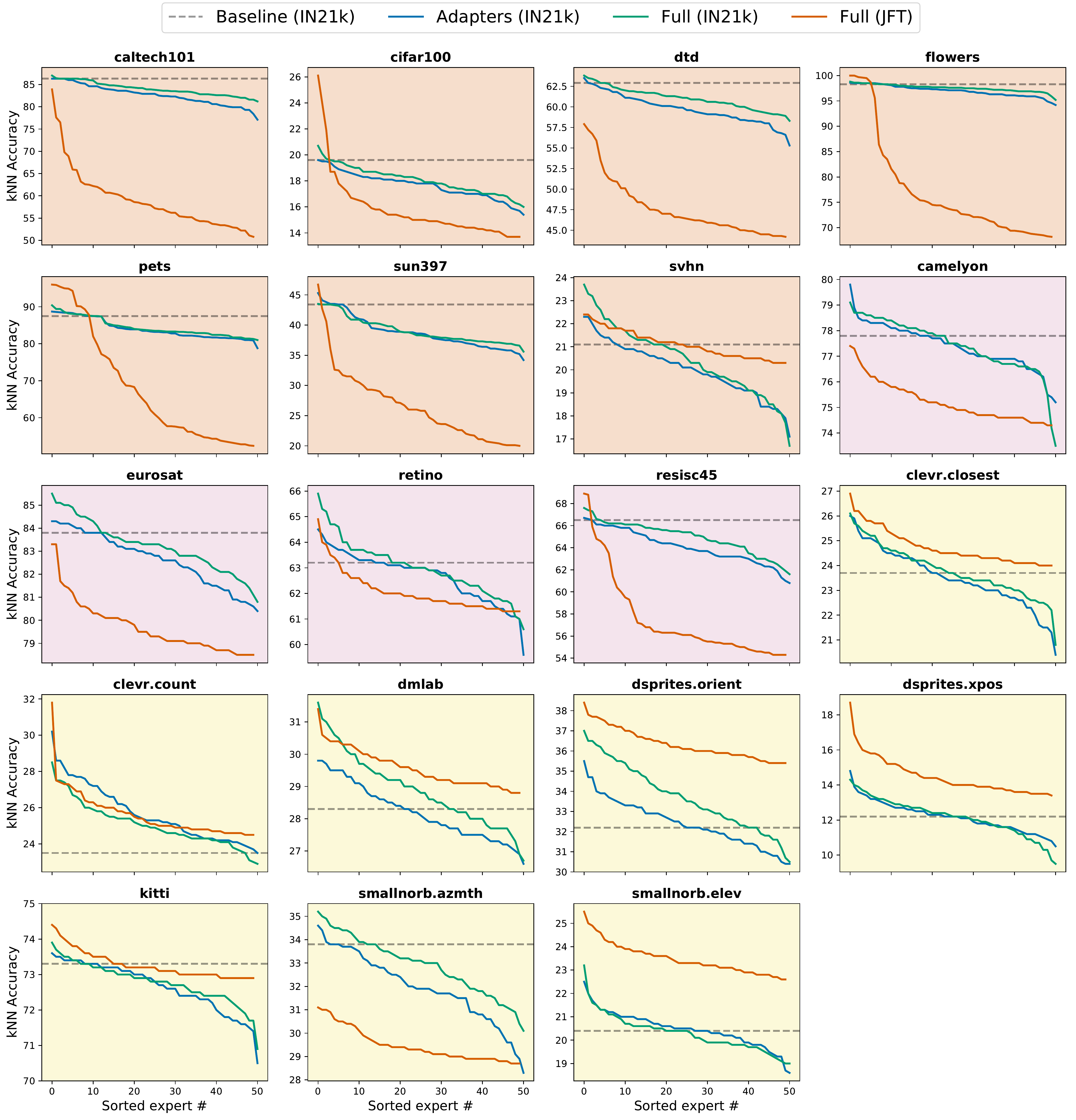}
\caption{%
Distribution of the kNN accuracy from experts trained on ImageNet21k. 
The dashed lines shows the kNN accuracy of the baseline model. 
The performance of the top-50 JFT full experts on each dataset is also shown.
In each dataset, the experts are sorted according to their accuracy.
The background color of the plot represents the group of the dataset:
\naturalsym~\textsc{Natural}; 
\specializedsym~\textsc{Specialized}; 
\structuredsym~\textsc{Structured}.
\label{fig:knn_accuracy_inet_distribution}}
\end{figure}

\Cref{fig:knn_accuracy_inet_distribution} presents the distribution of the kNN 
accuracy obtained from the embedding of each of the experts trained on the ImageNet21k dataset.
For context, in all the plots we also show the Top-50 JFT full experts.

For {\naturalsym~\textsc{Natural}} tasks, we observe that the quality of the ImageNet21k expert 
representations is way more homogeneous than the JFT one.
Accordingly, finding the right expert in JFT may be more important (as accuracy decreases fast),
while it may provide even more target-tailored representations (see Oxford Flowers and Oxford Pets).
Overall, ImageNet21k accuracies seem more stable, and differences between full and adapters are modest.

Overall, both full and adapter ImageNet21k experts seem to perform similarly on 
{\specializedsym~\textsc{Specialized}} tasks.
The plots suggest that ImageNet21k experts are a bit ahead of the full JFT ones 
(even though the gap at the top tends to close).

We see a few distinct behaviors in {\structuredsym~\textsc{Structured}}.
There tend not to be very remarkable winner experts, and full experts may provide a small boost compared to adapter-based ones.
In most datasets, the Top-50 full JFT experts outperform the ImageNet21k ones.

\clearpage
\subsection{kNN Accuracy Distribution for Consecutive Checkpoints of ImageNet21k Baseline}

\begin{figure}[htb]
\centering
\includegraphics[width=\textwidth]{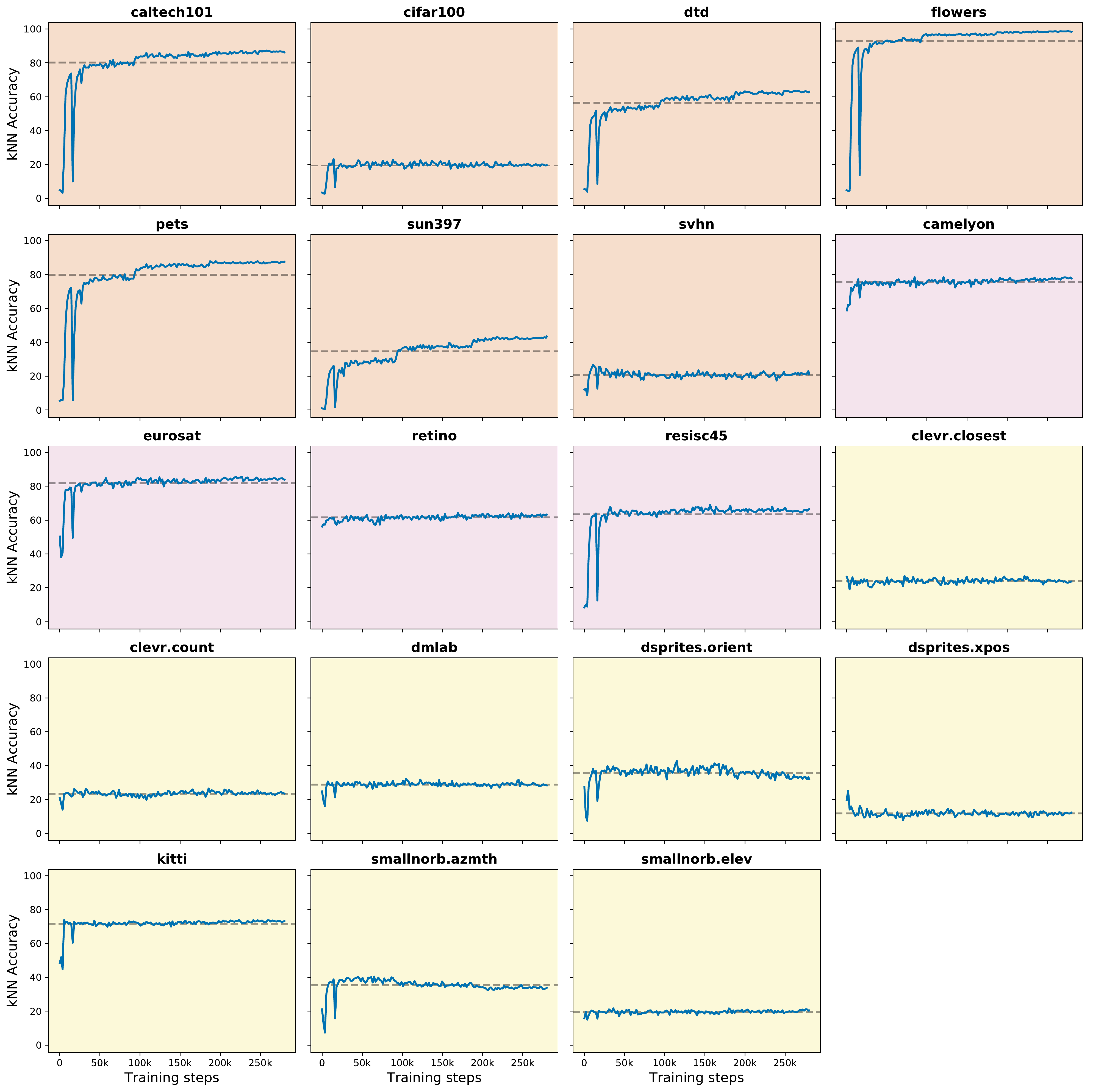}
\caption{
Accuracy of kNN using consecutive checkpoints stored during the ImageNet21k baseline 
training (90 epochs). The dashed line represents
the mean value across all checkpoints and it 
is useful to point out lack of improvement over 
time in some cases. The different types of 
datasets are highlighted by the background color:
{\naturalsym~\textsc{Natural}},
{\specializedsym~\textsc{Specialized}} and
{\structuredsym~\textsc{Structured}}.
\label{fig:app_inet_knn_ckpts}}
\end{figure}

In this subsection, we study how representations evolve during training.
In order to do that, we stored 157 checkpoints --equally spaced-- over the training of our ImageNet21k baseline.
We trained the model for 90 epochs.
For each dataset, we compute the kNN accuracy of the checkpoints, and display the curves in \cref{fig:app_inet_knn_ckpts}.
As an auxiliary line, we also show the mean kNN accuracy across all the checkpoints.

There are some clear differences depending on the \emph{type} of dataset.
In the case of {\naturalsym~\textsc{Natural}} images, it seems that more training leads to 
better representations.
The kNN accuracy tends to increase (Cifar100 and SVHN are exceptions).
{\specializedsym~\textsc{Specialized}} datasets behave in a different way; while there 
is an initial boost 
in accuracy (i.e.\ trained models are better than randomly initialized ones), long training
only leads to very minor improvements in representation quality for these tasks.
Finally, {\structuredsym~\textsc{Structured}} datasets have extremely flat footprints.
This probably means that our semantic experts are not a good fit for this type of task.

\subsection{Selected Experts}
\label{app:selected_experts}

The following table presents the experts selected by kNN in 
each of the individual datasets of the 
{\naturalsym~\textsc{Natural}},
{\specializedsym~\textsc{Specialized}} and 
{\structuredsym~\textsc{Structured}} groups.
\begin{table}[htb]
\centering
\caption{%
Selected experts by kNN using different expert architectures, 
in each of the VTAB-1k datasets. Datasets are grouped by
{\naturalsym~\textsc{Natural}},
{\specializedsym~\textsc{Specialized}} and 
{\structuredsym~\textsc{Structured}}.
\label{tab:knn_selected_experts}}
\resizebox{\textwidth}{!}{%
\begin{tabular}{ lcccc } 
\toprule
\textbf{Dataset} &
\textbf{Full-JFT} &
\textbf{Adapter-JFT} &
\textbf{Full-INet} &
\textbf{Adapter-INet} \\
\midrule
\naturalsym~Caltech101 & \emph{Baseline} & \emph{Baseline} & Physical Entity & \emph{Baseline} \\
\naturalsym~Cifar100 & \emph{Baseline} & \emph{Baseline} & Object & \emph{Baseline} \\
\naturalsym~DTD & \emph{Baseline} & \emph{Baseline} & Artifact & Artifact \\
\naturalsym~Oxford Flowers & Plant & Flower & Organism & Physical Entity \\
\naturalsym~Oxford Pets & Mammal & Carnivore & Carnivore & Carnivore \\
\naturalsym~Sun397 & Structure & \emph{Baseline} & Structure & Structure \\
\naturalsym~SVHN & Textile & Staple food & Implement & Artifact \\
\specializedsym~Diabetic Retinopathy  & Paper & Material & Food & Plant \\
\specializedsym~Eurosat & Snow & \emph{Baseline} & Whole & Breathe \\
\specializedsym~Patch Camelyon & Tree & \emph{Baseline} & Whole & Woody Plant \\
\specializedsym~Resisc45 & Geographical feature & \emph{Baseline} & Instrument & Arthropod \\
\structuredsym~Clevr Closest & Mode of transport & Canis & Mammal & Relation \\
\structuredsym~Clevr Count & Snow & Adventure & Spermatophyte & Flower \\
\structuredsym~DMLab & Sports equipment & Art & Implement & Vertebrate \\
\structuredsym~dSprites Orientation & Flowering plant & Mode of transport & Clothing & Implement \\
\structuredsym~dSprites Position & Dish & Toyota & Matter & Chordate \\
\structuredsym~Kitti & Shoe & Geographical feature & Plant & Abstraction \\
\structuredsym~Smallnorb Azimuth & Home and garden & Food & Abstraction & Device \\
\structuredsym~Smallnorb Elevation & Bag & Artwork & Carnivore & Mammal \\
\bottomrule
\end{tabular}%
}%
\end{table}

\section{Upstream training}
\label{sec:upstream_training_details}

\subsection{Upstream Training Details}
\label{subsec:exp_upstream_training}

\textbf{Unconditional pre-training}.
We pre-train generic \JFT and ImageNet21k models using 
a similar protocol to the one described in \cite{kolesnikov2019large}. 
In particular, we use SGD with momentum of 0.9, with a batch size of
4096, an initial learning rate of 0.03 (scaled by a factor of 
$\frac{\text{batch size}}{256}$), and weight decay of 0.001. 
The \JFT backbone model is trained for a total of 30 epochs, while 
the ImageNet21k is trained for 90 epochs. 
In both cases we perform training warm-up during the first 5\,000 
steps by linearly increasing learning rate, and then decay the
learning rate by a factor of 10 at
$\{\frac{1}{3},\frac{2}{3},\frac{5}{6}\}$ of the total duration.
During this phase we used a Cloud TPUv3-512 to train each of the
baseline models, which takes about 25 hours in the case of \JFT 
and .

\textbf{Experts training}.
In order to obtain the expert models, we then further tune these 
baselines (adding the residual adapters, when applicable) on different
subsets of the original upstream dataset. We use a similar
setting to the one described before, although we train for much 
shorter times, use a batch size of 1\,024, and use different 
learning rates. In particular, 
the full experts use an initial learning rate $10^{-4}$, since
all the parameters were pre-trained in the earlier phase. 
The experts with adapters use a larger learning rate of $10^{-1}$, 
as these components are trained from scratch and are the only ones
that are tuned. We use the same learning scaling factor and decay 
schedule as in the previous step. The initial learning rate in each 
case was decided based on average upstream performance across the
different expert datasets.
We fine-tune the full experts for 2 epochs, and the adapters 
for 4 epochs, relative to the size of the entire dataset. The only
exception is for the results reported in the comparison with 
\cite{ngiam2018domain}, for which we observed in the validation data
that full experts trained for 4 epochs performed better.

In both stages, we perform standard data augmentation during training,
which includes random image cropping as in \cite{Szegedy_2015_CVPR}, 
random horizontal mirroring, and finally image resize to a fixed 
resolution of $224 \times 224$. When we need to evaluate these models
on upstream data (i.e. upstream learning rate selection), 
we simply resize and crop the images to a 
fixed resolution of $224 \times 224$. 
Pixel values are converted to the $[-1, 1]$ range.

During this phase we used a Cloud TPUv3-32 to train each of the
experts. Training one of the \JFT experts for 2 epochs takes about 
11 hours, while this is reduced to 30 minutes in the case of the 
ImageNet21k experts.

\subsection{Upstream Freezing}

Note that many expert datasets $\mathbf{D}_e$ do not contain any instances of some original upstream classes (for example, the data for the expert \emph{elephant} may not contain any image with the label \emph{vehicle}).
As the head is frozen and shared among all experts, the adapters need to find other ways to ignore classes that do not apply at all to the expert.
We found this to be beneficial in practice, as we avoided too much upstream dependence on the head (which is later discarded in the transfer stage).
\section{Visual Task Adaptation Benchmark details}
\label{sec:vtab_details}

\subsection{Classes per Task}
\label{sec:vtab_data_details}
\begin{figure}[htb]
    \centering
  \includegraphics[width=0.7\linewidth]{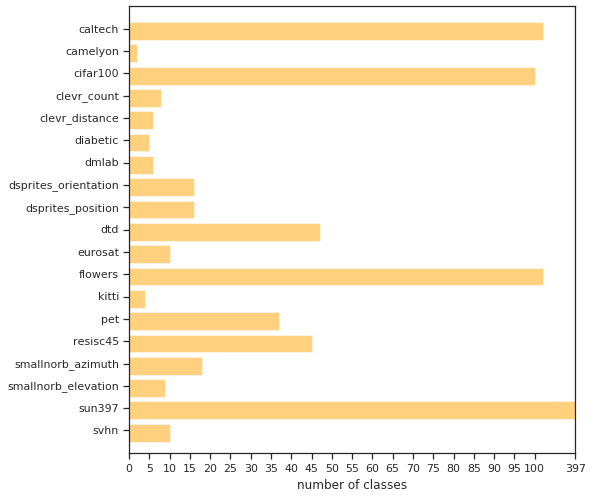}
  \caption{Number of classes per Downstream Task in VTAB.}
  \label{fig:vtab_classes}
\end{figure}

The number of classes in the VTAB tasks varies significantly, see \Cref{fig:vtab_classes}. 
As we are most interested in the low-data regime, we fix the number of downstream examples to 1\,000, implying that some downstream datasets only contain 3-10 examples per class --like Sun397 or Caltech.

\subsection{Downstream Hyperparameters on VTAB}
\label{subsec:vtab_hyperparams}

We use SGD with momentum of 0.9 and a batch size of 512. 
The initial learning rate is scaled by $\frac{\text{batch size}}{256}$. 
We don't perform any data augmentation, and resize all the images
to a fixed resolution of $224 \times 224$.
Pixel values are converted to the $[-1, 1]$ range.
We perform a restricted hyperparameter search, in particular 
we follow the \emph{lightweight} suggestion from \cite{zhai2019visual}.
The sweep tries in total 4 different sets of hyperparameters for
each dataset.
\begin{itemize}
\item Initial learning rate: the values of $\{0.1, 0.01\}$.
\item Training schedule: we try with a total training duration
of ${2\,500, 10\,000}$ steps, with a linear warm up of the scaling
rate of 200 and 500 steps, respectively.
For both durations, the learning rate is reduced by a factor of 10 
after $\{\frac{1}{3}, \frac{2}{3}, \frac{9}{10}\}$ of the
total number of steps.
\end{itemize}

Note that the hyperparameter sweep is done by training the models 
on 800 examples (out of the 1\,000 available), 
and selecting over 200 validation examples.
Then, the best combination of hyperparameters is used to re-train 
the models on the full 1\,000 data points.

Because the variability due to random initialization with so few 
data points is large in some datasets, we perform 10 independent 
runs of hyperparameter selection, and then re-training the models 
3 times for each of the 10 selected hyperparameters. 
This yields a total of 30 outcomes for each of the VTAB-1k datasets.
For each dataset, we report the median over these 30 trials and 
compute (percentile) bootstrapped confidence intervals at 95\% level.

For downstream training we use a Cloud TPUv3-16. In VTAB-1k, the running time 
depends on the number of steps. It takes 12 minutes to fine-tune one of our 
experts for 10\,000 steps (the longest duration), and just 3 minutes for the 
shorter schedule of 2\,500 steps.

\subsection{Additional Results of Different Expert Selection Strategies}
\label{subsec:results_selection_strategies_extra}

In \cref{subsec:experiments_selection_algorithm}, and more precisely in \cref{tab:vtab_test_transfer_algorithms}, we studied the performance of random transfer (expert selection uniformly at random).
There, we only reported the results corresponding to full experts trained on JFT.
For completeness, \cref{tab:selection_algorithms_full_and_adapters} shows also the results with adapters.
The pattern is fairly similar: random transfer leads to massive losses in Natural datasets (we obtain an almost 35\% improvement by applying kNN with respect to random transfer).
Domain Prediction and Label Matching also heavily help in these settings.
In Specialized and Structured tasks, both Domain Prediction and Label Matching seem to offer little-to-no gains, whereas kNN still leads to a modest boost on Structured, and a decent one on Specialized.

Overall (last column of the table), the improvement is significant and strong for all routing methods.

\begin{table}[htb]
\centering
\caption{VTAB-1k results of different selection algorithms, using full and adapter experts
trained on \JFT. The average accuracy across each group of tasks and
across all VTAB is reported. In each dataset, the median accuracy over 30 runs 
is used. Bootstrapped confidence intervals at 95\% level.}
\label{tab:selection_algorithms_full_and_adapters}
\begin{tabular}{lllll}
\toprule
& \textsc{Natural} & \textsc{Specialized} & \textsc{Structured} & \textsc{All}\\
\midrule  
\emph{Adapters} & & & &\\
~~Random
  & \cih{58.6}{56.1}{59.6} 
  & \cih{78.3}{76.8}{79.2} 
  & \cih{58.6}{57.8}{59.6} 
  & \cih{62.8}{61.7}{63.3} \\
~~Domain Prediction
  & \cih{70.8}{69.3}{71.6} 
  & \cih{75.5}{63.7}{78.0} 
  & \cih{59.7}{58.2}{61.0} 
  & \cih{67.1}{64.5}{67.9} \\
~~Label Matching
  & \cih{75.3}{75.1}{75.4} 
  & \cih{80.5}{78.2}{81.3} 
  & \cih{56.1}{51.8}{57.0} 
  & \cih{68.3}{66.4}{68.7} \\
~~Performance Proxy
  & \cih{79.0}{78.6}{79.1} 
  & \cih{81.3}{79.2}{82.5} 
  & \cih{59.1}{58.3}{60.1} 
  & \cih{71.1}{70.5}{71.6} \\
\midrule
\emph{Full} & & & &\\
~~Random
  & \cih{60.6}{59.1}{63.9} 
  & \cih{81.22}{80.9}{81.8} 
  & \cih{56.8}{54.9}{57.8} 
  & \cih{63.3}{62.3}{64.6} \\
~~Domain Prediction
  & \cih{75.9}{74.4}{77.4} 
  & \cih{81.5}{81.3}{82.2} 
  & \cih{\textbf{57.0}}{56.1}{57.4} 
  & \cih{69.1}{68.4}{69.8} \\
~~Label Matching
  & \cih{78.0}{77.8}{78.1} 
  & \cih{80.3}{79.1}{82.5} 
  & \cih{56.9}{55.6}{57.2} 
  & \cih{69.6}{68.9}{70.0} \\
~~Performance Proxy
  & \cih{\textbf{79.7}}{79.5}{80.0} 
  & \cih{\textbf{83.6}}{83.3}{83.8} 
  & \cih{55.3}{52.1}{56.3} 
  & \cih{\textbf{70.2}}{68.9}{70.6} \\
\bottomrule     
\end{tabular}
\end{table}

\subsection{Per-Task Results}
\label{subsec:vtab_per_task_results}
\begin{figure}[htb]
\centering
\includegraphics[width=\textwidth]{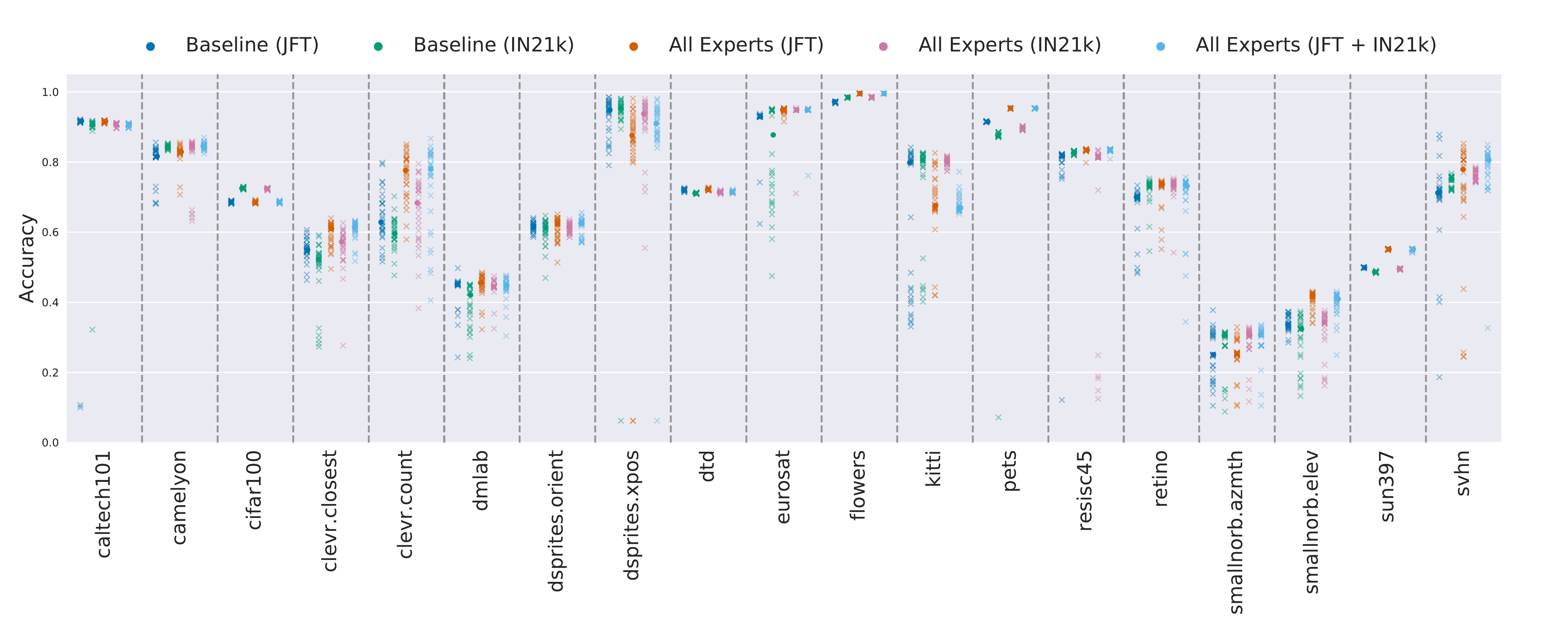}
\caption{%
VTAB-1k accuracy in all datasets for 30 runs using different baselines and experts models trained 
on \JFT and ImageNet21k. The median is represented by a darker
point. Best seen in color.%
\label{fig:vtab_per_dataset_all_runs}}
\end{figure}

All VTAB results presented so far were averaged over dataset types (natural, specialized, and structured).
In this subsection, we break down the outcomes per dataset.
\Cref{tab:vtab_per_dataset_all} shows the mean accuracy (and confidence intervals) for 13 algorithms and 19 datasets.
The datasets are sorted according to the data type.
The table can be used for reference.
Some datasets showcase a wide range of outcomes for any fixed algorithm.
In order to expose this in a clear fashion, we present in \cref{fig:vtab_per_dataset_all_runs} the individual trial accuracies for the best algorithms and baselines, in all of the VTAB datasets.
The fine-tuning process on datasets like Clver Count, dSprites xPosition, or SVHN definitely shows a high variance of test accuracies.
The median estimator partially mitigates this effect.

\newpage
\begin{landscape}
\begin{table}[tb]
\centering
\caption{%
Accuracy on the individual datasets of the VTAB-1k 
benchmark.
Algorithms include experts trained on both JFT and ImageNet21k, with adapters and full
architectures, and by means of different selection methods 
(Expert Predictor Network, EPN; 
Kullback--Leibler, KL; and kNN). 
In each dataset, the median accuracy over 30 runs is used.
Bootstrapped confidence intervals at 95\% are shown. Color indicates dataset group:
{\naturalsym~\textsc{Natural}};
{\specializedsym~\textsc{Specialized}};
{\structuredsym~\textsc{Structured}}.
\label{tab:vtab_per_dataset_all}}
{\setlength{\tabcolsep}{2pt}
\begin{tabular}{lrrrrrrrrrrrrrrrrrrr}
\toprule
& \rotatebox[origin=l]{90}{\naturalsym~caltech101}
& \rotatebox[origin=l]{90}{\naturalsym~cifar100}
& \rotatebox[origin=l]{90}{\naturalsym~dtd}
& \rotatebox[origin=l]{90}{\naturalsym~flowers}
& \rotatebox[origin=l]{90}{\naturalsym~pets}
& \rotatebox[origin=l]{90}{\naturalsym~sun397}
& \rotatebox[origin=l]{90}{\naturalsym~svhn}
& \rotatebox[origin=l]{90}{\specializedsym~camelyon}
& \rotatebox[origin=l]{90}{\specializedsym~eurosat}
& \rotatebox[origin=l]{90}{\specializedsym~retino}
& \rotatebox[origin=l]{90}{\specializedsym~resisc45}
& \rotatebox[origin=l]{90}{\structuredsym~clevr.closest}
& \rotatebox[origin=l]{90}{\structuredsym~clevr.count}
& \rotatebox[origin=l]{90}{\structuredsym~dmlab}
& \rotatebox[origin=l]{90}{\structuredsym~dsprites.orient}
& \rotatebox[origin=l]{90}{\structuredsym~dsprites.xpos}
& \rotatebox[origin=l]{90}{\structuredsym~kitti}
& \rotatebox[origin=l]{90}{\structuredsym~smallnorb.azmth}
& \rotatebox[origin=l]{90}{\structuredsym~smallnorb.elev}
\\
\midrule
\emph{JFT} & & & & & & & & & & & & & & & & & & & \\
\cit{~~Baseline}
& \civ{91.7}{91.5}{91.8} 
& \civ{68.6}{68.3}{68.7} 
& \civ{72.1}{72.0}{72.2} 
& \civ{97.2}{97.1}{97.2} 
& \civ{91.5}{91.4}{91.5} 
& \civ{49.9}{49.9}{50.0} 
& \civ{71.2}{70.5}{72.0} 
& \civ{81.6}{81.4}{83.1} 
& \civ{93.0}{92.9}{93.1} 
& \civ{70.0}{69.6}{70.2} 
& \civ{81.8}{81.5}{81.9} 
& \civ{54.9}{54.5}{55.8} 
& \civ{62.8}{60.7}{68.0} 
& \civ{45.1}{45.0}{45.3} 
& \civ{61.6}{61.1}{62.1} 
& \civ{94.9}{93.7}{96.2} 
& \civ{79.8}{43.9}{80.8} 
& \civ{25.1}{22.0}{30.5} 
& \civ{33.6}{33.1}{35.5} 
\\
\cit{~~Adapters (EPN)}
& \civ{91.7}{91.6}{91.7} 
& \civ{34.0}{32.2}{34.3} 
& \civ{58.3}{58.0}{58.6} 
& \civ{98.2}{98.1}{98.2} 
& \civ{91.4}{91.3}{91.5} 
& \civ{48.3}{48.1}{48.4} 
& \civ{73.7}{63.3}{79.2} 
& \civ{79.6}{71.7}{83.1} 
& \civ{93.1}{92.2}{93.6} 
& \civ{60.0}{53.7}{66.4} 
& \civ{69.3}{23.3}{74.3} 
& \civ{60.6}{57.3}{61.9} 
& \civ{63.3}{56.7}{73.3} 
& \civ{45.1}{43.7}{46.0} 
& \civ{59.3}{56.9}{59.5} 
& \civ{95.8}{95.0}{96.8} 
& \civ{79.2}{73.7}{80.4} 
& \civ{32.6}{32.1}{33.1} 
& \civ{41.8}{37.7}{42.6} 
\\
\cit{~~Adapters (KL)}
& \civ{91.4}{91.1}{91.6} 
& \civ{68.3}{68.2}{68.3} 
& \civ{57.5}{57.0}{57.7} 
& \civ{98.1}{97.8}{98.2} 
& \civ{92.0}{91.9}{92.0} 
& \civ{48.3}{48.3}{48.4} 
& \civ{71.6}{70.9}{72.1} 
& \civ{83.0}{74.1}{83.6} 
& \civ{93.0}{92.9}{93.1} 
& \civ{68.2}{64.5}{70.7} 
& \civ{77.7}{77.5}{79.7} 
& \civ{52.2}{49.3}{53.8} 
& \civ{59.9}{58.3}{62.3} 
& \civ{43.8}{41.7}{45.0} 
& \civ{61.2}{59.9}{62.1} 
& \civ{94.1}{90.8}{96.1} 
& \civ{77.9}{47.0}{79.0} 
& \civ{24.9}{16.8}{30.4} 
& \civ{34.4}{33.4}{35.4} 
\\
\cit{~~Adapters (kNN)}
& \civ{91.6}{91.5}{91.7} 
& \civ{68.4}{68.3}{68.6} 
& \civ{72.2}{72.1}{72.2} 
& \civ{97.7}{97.7}{97.8} 
& \civ{91.9}{91.8}{92.0} 
& \civ{49.8}{49.8}{49.9} 
& \civ{81.1}{78.7}{81.8} 
& \civ{83.3}{82.8}{83.7} 
& \civ{93.0}{93.0}{93.1} 
& \civ{67.3}{58.8}{71.8} 
& \civ{81.7}{81.5}{81.8} 
& \civ{61.2}{59.1}{62.1} 
& \civ{78.0}{75.0}{79.5} 
& \civ{44.9}{44.3}{46.2} 
& \civ{61.8}{61.3}{62.2} 
& \civ{89.3}{87.5}{92.8} 
& \civ{78.8}{77.6}{79.4} 
& \civ{24.1}{23.7}{26.6} 
& \civ{34.5}{30.3}{40.7} 
\\
\cit{~~Full (EPN)}
& \civ{91.6}{91.2}{91.7} 
& \civ{53.2}{53.1}{53.4} 
& \civ{63.3}{63.0}{63.6} 
& \civ{99.5}{99.5}{99.5} 
& \civ{96.1}{96.1}{96.2} 
& \civ{55.1}{55.0}{55.1} 
& \civ{72.3}{61.7}{82.6} 
& \civ{83.3}{82.9}{84.1} 
& \civ{94.0}{93.9}{94.1} 
& \civ{74.1}{73.9}{74.2} 
& \civ{74.5}{74.3}{77.1} 
& \civ{56.8}{55.2}{57.9} 
& \civ{62.9}{60.6}{65.5} 
& \civ{38.2}{37.4}{39.5} 
& \civ{64.2}{63.1}{64.5} 
& \civ{96.5}{95.5}{97.0} 
& \civ{75.5}{74.4}{76.1} 
& \civ{22.8}{19.1}{23.3} 
& \civ{39.1}{38.7}{39.7} 
\\
\cit{~~Full (KL)}
& \civ{91.6}{91.5}{91.7} 
& \civ{68.4}{68.2}{68.6} 
& \civ{64.3}{63.8}{64.7} 
& \civ{99.5}{99.5}{99.5} 
& \civ{95.9}{95.9}{95.9} 
& \civ{55.1}{55.1}{55.2} 
& \civ{70.7}{70.2}{71.8} 
& \civ{83.1}{81.8}{83.5} 
& \civ{93.0}{92.9}{93.0} 
& \civ{61.8}{57.4}{70.4} 
& \civ{83.4}{83.4}{83.5} 
& \civ{54.0}{50.3}{55.3} 
& \civ{60.6}{56.5}{61.3} 
& \civ{45.1}{41.5}{45.2} 
& \civ{64.5}{62.9}{65.2} 
& \civ{97.3}{96.8}{97.5} 
& \civ{75.0}{74.8}{75.4} 
& \civ{24.9}{18.3}{25.2} 
& \civ{34.0}{32.0}{35.2} 
\\
\cit{~~Full (kNN)}
& \civ{91.4}{91.2}{91.6} 
& \civ{68.7}{68.4}{68.9} 
& \civ{72.2}{72.0}{72.3} 
& \civ{99.5}{99.5}{99.5} 
& \civ{95.4}{95.3}{95.4} 
& \civ{55.1}{55.1}{55.2} 
& \civ{75.3}{74.4}{77.4} 
& \civ{82.8}{82.5}{83.4} 
& \civ{94.8}{94.4}{95.1} 
& \civ{73.1}{72.1}{73.8} 
& \civ{83.4}{83.4}{83.5} 
& \civ{57.3}{55.3}{58.1} 
& \civ{54.4}{51.0}{57.7} 
& \civ{42.2}{41.8}{43.0} 
& \civ{60.6}{58.6}{62.2} 
& \civ{93.7}{88.9}{95.7} 
& \civ{67.0}{44.0}{71.3} 
& \civ{25.5}{25.3}{29.0} 
& \civ{41.5}{39.9}{42.2} 
\\
\cit{~~All Experts (kNN)}
& \civ{91.6}{91.4}{91.7} 
& \civ{68.5}{68.4}{68.6} 
& \civ{72.1}{72.0}{72.3} 
& \civ{99.5}{99.5}{99.5} 
& \civ{95.4}{95.3}{95.4} 
& \civ{55.1}{55.1}{55.2} 
& \civ{77.9}{71.8}{80.7} 
& \civ{82.9}{82.6}{83.2} 
& \civ{94.9}{94.7}{95.0} 
& \civ{73.7}{73.4}{73.9} 
& \civ{83.4}{83.3}{83.5} 
& \civ{61.0}{58.4}{61.8} 
& \civ{77.6}{74.2}{80.7} 
& \civ{45.6}{44.8}{46.4} 
& \civ{62.3}{60.0}{62.9} 
& \civ{87.6}{85.7}{90.6} 
& \civ{67.6}{66.8}{71.8} 
& \civ{25.3}{25.1}{25.6} 
& \civ{41.8}{40.9}{42.1} 
\\
\midrule
\emph{IN21k} & & & & & & & & & & & & & & & & & & & \\
\cit{~~Baseline}
& \civ{90.8}{90.7}{91.0} 
& \civ{72.5}{72.4}{72.6} 
& \civ{71.1}{71.0}{71.2} 
& \civ{98.5}{98.4}{98.5} 
& \civ{87.6}{87.4}{87.7} 
& \civ{48.6}{48.6}{48.7} 
& \civ{74.9}{72.6}{75.5} 
& \civ{84.3}{84.1}{84.5} 
& \civ{87.7}{73.4}{94.6} 
& \civ{73.2}{72.9}{73.8} 
& \civ{82.8}{82.2}{83.0} 
& \civ{52.2}{51.1}{53.4} 
& \civ{59.6}{58.1}{61.8} 
& \civ{42.1}{37.5}{43.2} 
& \civ{61.3}{60.0}{62.2} 
& \civ{95.4}{94.4}{96.0} 
& \civ{80.5}{78.7}{81.6} 
& \civ{30.6}{27.7}{30.7} 
& \civ{32.4}{29.7}{33.3} 
\\
\cit{~~Adapters (kNN)}
& \civ{89.9}{89.7}{90.9} 
& \civ{72.4}{72.3}{72.6} 
& \civ{71.2}{71.1}{71.3} 
& \civ{98.4}{98.4}{98.4} 
& \civ{89.4}{89.3}{89.6} 
& \civ{49.5}{49.4}{49.6} 
& \civ{75.8}{75.3}{76.2} 
& \civ{83.9}{83.1}{84.3} 
& \civ{94.6}{94.5}{94.6} 
& \civ{73.8}{73.2}{74.0} 
& \civ{81.8}{80.5}{82.0} 
& \civ{54.9}{53.5}{55.5} 
& \civ{64.0}{61.1}{68.4} 
& \civ{44.9}{44.2}{45.2} 
& \civ{61.3}{60.2}{61.9} 
& \civ{93.6}{91.4}{94.3} 
& \civ{78.6}{77.0}{79.5} 
& \civ{27.8}{27.7}{31.2} 
& \civ{34.9}{34.4}{35.6} 
\\
\cit{~~Full (kNN)}
& \civ{90.7}{90.3}{90.9} 
& \civ{72.5}{72.5}{72.5} 
& \civ{71.4}{71.4}{71.5} 
& \civ{98.6}{98.5}{98.6} 
& \civ{89.9}{89.4}{90.1} 
& \civ{49.7}{49.6}{49.7} 
& \civ{74.9}{74.5}{77.2} 
& \civ{83.8}{83.5}{84.2} 
& \civ{94.9}{94.8}{94.9} 
& \civ{73.6}{72.7}{74.0} 
& \civ{81.5}{81.5}{81.7} 
& \civ{58.2}{55.6}{59.4} 
& \civ{68.9}{65.8}{70.6} 
& \civ{45.1}{44.5}{45.7} 
& \civ{60.9}{59.5}{62.0} 
& \civ{95.3}{93.5}{96.2} 
& \civ{80.4}{79.7}{80.9} 
& \civ{31.1}{30.5}{31.8} 
& \civ{35.2}{34.7}{35.8} 
\\
\cit{~~All Experts (kNN)}
& \civ{90.8}{90.7}{90.9} 
& \civ{72.4}{72.2}{72.5} 
& \civ{71.4}{71.3}{71.4} 
& \civ{98.5}{98.4}{98.6} 
& \civ{89.6}{89.4}{90.1} 
& \civ{49.5}{49.5}{49.6} 
& \civ{76.0}{74.7}{77.5} 
& \civ{84.5}{84.0}{84.8} 
& \civ{94.8}{94.8}{94.9} 
& \civ{73.5}{73.2}{74.1} 
& \civ{81.4}{81.3}{81.5} 
& \civ{57.3}{55.2}{58.5} 
& \civ{68.3}{62.6}{72.1} 
& \civ{44.4}{44.3}{44.6} 
& \civ{61.4}{60.4}{62.0} 
& \civ{93.8}{92.1}{95.1} 
& \civ{80.2}{79.6}{80.7} 
& \civ{30.6}{30.1}{31.3} 
& \civ{34.3}{33.9}{35.4} 
\\
\midrule
\emph{JFT + IN21k} & & & & & & & & & & & & & & & & & & \\
\cit{~~All Experts (kNN)}
& \civ{90.7}{90.6}{90.8} 
& \civ{68.5}{68.4}{68.7} 
& \civ{71.4}{71.3}{71.5} 
& \civ{99.5}{99.5}{99.5} 
& \civ{95.4}{95.3}{95.4} 
& \civ{55.2}{55.1}{55.2} 
& \civ{80.6}{78.1}{81.5} 
& \civ{84.5}{84.3}{84.8} 
& \civ{94.9}{94.8}{94.9} 
& \civ{73.2}{72.1}{73.9} 
& \civ{83.5}{83.4}{83.6} 
& \civ{61.4}{60.6}{62.2} 
& \civ{78.1}{73.2}{81.8} 
& \civ{44.8}{44.0}{45.4} 
& \civ{62.5}{61.8}{63.0} 
& \civ{91.0}{88.1}{93.6} 
& \civ{66.9}{66.5}{67.9} 
& \civ{30.8}{27.7}{31.2} 
& \civ{40.9}{39.4}{41.7} 
\\
\bottomrule
\end{tabular}}
\end{table}%
\end{landscape}

\section{Details on the Comparison with Domain Adaptive Transfer}
\label{sec:domain_adaptive_details}

\subsection{Hyperparameters}
\label{subsec:domain_adaptive_hyperparms}

We randomly explored the space of hyperparameters drawing 36 samples from the following distributions:
\begin{itemize}
    \item Initial learning rate: Log-uniform in $[2 \cdot 10^{-4}, 2 \cdot 10^{-1}]$. 
    \item Total training steps: Uniform in $\{2000, 4000, 8000, 16000, 32000\}$. 
    \item Weight decay: Log-uniform in $[10^{-6}, 10^{-2}]$.
    \item Mixup $\alpha$: Uniform in $\{0, 0.05, 0.1, 0.2, 0.4\}$.
\end{itemize}
We use a batch size equal to 512 and decay the learning rate by 0.1 at 30\%, 60\% and 90\% 
of the training duration. During the first 10\% of
training steps, we linearly warm up the learning rate.
Standard data augmentation techniques are applied to prevent overfitting. 
During training, for all datasets except CIFAR10 (which has a smaller resolution),
we resized the images to a fixed size of 512 pixels on both sides, then randomly cropped
it to a patch of 480 pixels, randomly flipped the image horizontally, and converted the 
pixel values to the $[-1, 1]$ range. For CIFAR10, we used a resolution of 160 pixels
during the resize and crops of 128 pixels. During evaluation, we simply resize the images
and convert the pixel values analogously.

We find the best hyperparameter for each dataset based on the accuracy on
the validation data, and then, applying the corresponding set of hyperparameters, 
the selected expert is fine-tuned again on the union of the training and validation
examples of the dataset. The accuracy on a test set is reported. We re-trained the models 
30 times using different random seeds. 

For downstream training we use a Cloud TPUv3-16. In DAT, the running time
depends on the number of steps selected as the hyperparameter and the resolution of
the images. At most, it takes 150 minutes to fine-tune one of our experts on one
downstream dataset, and at least it takes 8 minutes.

\subsection{Detailed results}

\Cref{tab:dat_results_details} shows the mean accuracy across those 30 trials and the 
95\% bootstrapped confidence intervals, for each of our experts and selection
algorithms, for each dataset as well as the average across the six datasets.

\begin{table}[htb]
\centering
\caption{Accuracy on the datasets used in~\cite{ngiam2018domain}, and the average accuracy
across the six of them. Bootstrapped confidence intervals at 95\% are shown below the 
accuracy, when available. The first two rows are Inception-v3 models, as reported in~\cite{ngiam2018domain}. The rest of the rows are produced by our own models, based on
Resnet-50-v2, grouped by expert selection method. The suffixes ``2e'' and ``4e'' denote that
the expert modules were trained for 2 or 4 epochs, respectively.}
\label{tab:dat_results_details}
\setlength{\tabcolsep}{2pt}
\resizebox{1.0\textwidth}{!}{%
\begin{tabular}{llllllll}
\toprule
    & Aircraft & Birds & Cars & \hspace{-1ex}CIFAR10 & Food & Pets* & Avg.\\
\midrule
Baseline
& \cih{91.4}{91.0}{91.7} 
& \cih{78.8}{78.0}{79.4} 
& \cih{95.6}{95.4}{95.7} 
& \cih{97.8}{97.7}{97.9} 
& \cih{91.3}{91.2}{91.5} 
& \cih{94.5}{94.4}{94.6} 
& \cih{91.6}{91.4}{91.7} \\
kNN & & & & & & &\\
~~Adapters, 4e
& \cih{92.5}{92.2}{92.8} 
& \cih{79.4}{78.7}{80.1} 
& \cih{95.9}{95.8}{96.0} 
& \cih{97.9}{97.8}{98.0} 
& \cih{91.6}{91.5}{91.7} 
& \cih{94.6}{94.4}{94.8} 
& \cih{92.0}{91.9}{92.1} \\
~~Full, 2e
& \cih{94.5}{94.2}{94.7} 
& \cih{83.5}{83.1}{83.9} 
& \cih{96.0}{95.8}{96.2} 
& \cih{97.9}{97.8}{98.0} 
& \cih{92.9}{92.8}{93.1} 
& \cih{96.8}{96.7}{96.9} 
& \cih{93.6}{93.5}{93.7} \\
~~Full, 4e
& \cih{\textbf{94.8}}{94.5}{95.1} 
& \cih{83.6}{83.1}{83.9} 
& \cih{96.1}{96.0}{96.3} 
& \cih{97.8}{97.7}{97.9} 
& \cih{93.1}{92.8}{93.2} 
& \cih{\textbf{97.0}}{96.9}{97.1} 
& \cih{93.7}{93.6}{93.8} \\
KL & & & & & & &\\
~~Adapters, 4e
& \cih{92.1}{91.8}{92.5} 
& \cih{80.0}{79.5}{80.4} 
& \cih{95.9}{95.8}{96.0} 
& \cih{97.9}{97.8}{98.0} 
& \cih{91.6}{91.5}{91.8} 
& \cih{94.5}{94.3}{94.7} 
& \cih{92.0}{91.9}{92.1} \\
~~Full, 2e
& \cih{94.6}{93.6}{95.0} 
& \cih{83.1}{82.6}{83.5} 
& \cih{96.1}{96.0}{96.2}
& \cih{97.9}{97.8}{98.1} 
& \cih{92.9}{92.9}{93.1} 
& \cih{96.6}{96.5}{96.7} 
& \cih{93.5}{93.4}{93.7} \\
~~Full, 4e
& \cih{94.4}{94.1}{94.7} 
& \cih{83.7}{83.3}{84.3} 
& \cih{\textbf{96.4}}{96.3}{96.4} 
& \cih{97.9}{97.8}{98.0} 
& \cih{93.1}{92.9}{93.3} 
& \cih{96.6}{96.6}{96.6} 
& \cih{93.7}{93.6}{93.8} \\
EPN & & & & & & &\\  
~~Adapters, 4e
& \cih{92.1}{91.7}{92.5} 
& \cih{79.7}{79.1}{80.2} 
& \cih{95.8}{95.6}{96.0} 
& \cih{97.3}{97.1}{97.4} 
& \cih{91.5}{91.3}{91.7} 
& \cih{93.1}{91.8}{93.6} 
& \cih{91.6}{91.4}{91.8} \\
~~Full, 2e
& \cih{94.0}{93.6}{94.3} 
& \cih{83.3}{82.7}{83.9}
& \cih{96.2}{96.1}{96.2} 
& \cih{96.9}{96.7}{97.1} 
& \cih{92.5}{92.3}{92.6} 
& \cih{\textbf{97.0}}{97.0}{97.1} 
& \cih{93.3}{93.2}{93.4} \\
~~Full, 4e
& \cih{94.2}{94.1}{94.4}
& \cih{84.3}{83.9}{84.7} 
& \cih{96.0}{96.0}{96.1} 
& \cih{96.6}{96.4}{96.7} 
& \cih{92.4}{92.3}{92.6} 
& \cih{96.9}{96.9}{97.0} 
& \cih{93.4}{93.3}{93.5} \\
\midrule
Dom-Ad (In-v3) \cite{ngiam2018domain}
  & 94.1 & 81.7 & 95.7 & {98.3} & {94.1} & \textbf{97.1} & 93.5 \\ 
Dom-Ad (Am-B) \cite{ngiam2018domain}
  & 92.8 & \textbf{85.1} & 95.8 & \textbf{98.6} & \textbf{95.3} & {96.8} & \textbf{94.1} \\ 
\bottomrule
\end{tabular}
}
\begin{flushleft}
\small{*Pets results are mean per class accuracy as opposed to mean accuracy.}\end{flushleft}\vspace{-0.5cm}
\end{table}

\subsection{Differences in asymptotic running time}
\label{subsec:domain_adaptive_transfer_tradeoff}

\begin{table}[htb]
\caption{Asymptotic running times of Domain Adaptive Transfer (DAT) 
\cite{ngiam2018domain} 
and our work, where $P$ is the number of parameters of the network,
$B$ is the batch size, $S_U$ is the number of training steps of the baseline model, 
$S_A$ is the number of training steps for adapting the baseline model, 
$S_F$ is the number of training steps for fine-tuning the specialist model to 
the downstream task, and $E$ is the number of pre-trained experts in our approach.}
\label{tab:asymptotic_time_domain_adaptive}
\centering
\begin{tabular}{lrr}
\toprule
& DAT \cite{ngiam2018domain} & Ours\\
\midrule
Upstream training 
    & $O(S_U \cdot B \cdot P)$ 
    & $O((S_U + S_A \cdot E) \cdot B \cdot P)$ \\
Downstream & & \\    
~~Expert preparation 
    & $O((N_T + S_A \cdot B) \cdot P)$
    & $O((N_T \cdot P + N_T^2) \cdot E)$ \\
~~Fine-tuning 
    & $O(S_F \cdot B \cdot P)$
    & $O(S_F \cdot B \cdot P)$ \\
\bottomrule
\end{tabular}
\end{table}

\Cref{tab:asymptotic_time_domain_adaptive} contains an asymptotic 
analysis of the different phases of each approach. In our case,
upstream training includes both the cost of training the generic 
backbone network for $S_U$ steps, 
and the cost of training each of the $E$ experts for $S_A$ steps,
with a batch size of $B$.
In the case of Domain Adaptive Transfer (DAT), only the first cost
is incurred in this phase.
Observe that in our case the cost of upstream training is 
amortized over the number of tasks that one has to learn, 
since it's only incurred once. 

In the downstream phase, both methods need a forward pass over the
number of downstream examples, $N_T$. Then, leaving the number 
of parameters of the model aside, the cost of \cite{ngiam2018domain} 
is dominated by $S_A \cdot B$, which is the total number of examples 
used to fine-tune the baseline model to a weighted/resampled version 
of the upstream data, and ours is dominated by $N_T \cdot E$, the cost 
of running a forward pass of the downstream data in each of the 
pre-trained experts.
In practice, because $S_A \cdot B$ (roughly $1.2 \cdot 10^9$, when 
fine-tuning for 4 epochs on \JFT) is much larger than $N_T \cdot E$ 
(roughly $5 \cdot 10^5$, when using 1\,000 downstream examples for 
selecting over 500 experts), our approach is much faster.
Thus, our approach should be roughly three orders of magnitude 
faster than that of \cite{ngiam2018domain}, when learning a new task.
The final cost of fine-tuning to the downstream dataset is the same
in both cases, and it's negligible in comparison.

In \cref{subsec:adaptive_transfer_comparison} we actually measured 
the difference between selecting among 240 experts (R50) and 
fine-tuning the baseline model for 4 epochs on \JFT, using the same
hardware, and the difference was of $900\times$, so we estimate 
the real difference to be in the range ${500\times}-{1000\times}$, 
depending on implementation details.
\section{The Value of Semantic Experts}
\label{app:random_experts}

In \cref{subsec:experiments_selection_algorithm}, we have seen that a smart choice of experts leads to substantial gains with respect to 
a single model trained on all the upstream data. 
Here, we rule out the possibility that these gains come merely from the fact that we are able 
to select a representation among a wide range of choices by \emph{directly} testing their 
initial predictive power on the downstream task.
To do so, instead of training our experts in subsets of \JFT based on its label hierarchy,
we fully fine-tuned the baseline model on 240 uniformly random subsets of \JFT, 
with sizes matching the size of our original semantic experts.
We did this independently with both adapter and full experts.
Then, we applied kNN to select the best random expert on each downstream dataset.
Note this is not at all equivalent to applying transfer at random as in \cref{subsec:experiments_selection_algorithm}.

\begin{figure}[htb]
\centering
\subcaptionbox{JFT Adapter Experts}{%
\includegraphics[width=\textwidth]{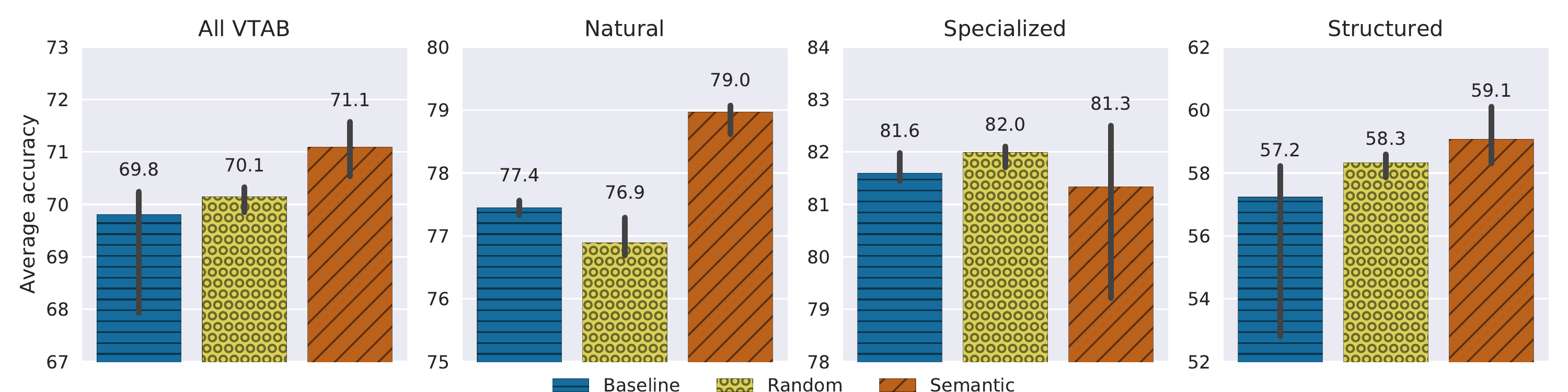}}
\subcaptionbox{JFT Full Experts}{%
\includegraphics[width=\textwidth]{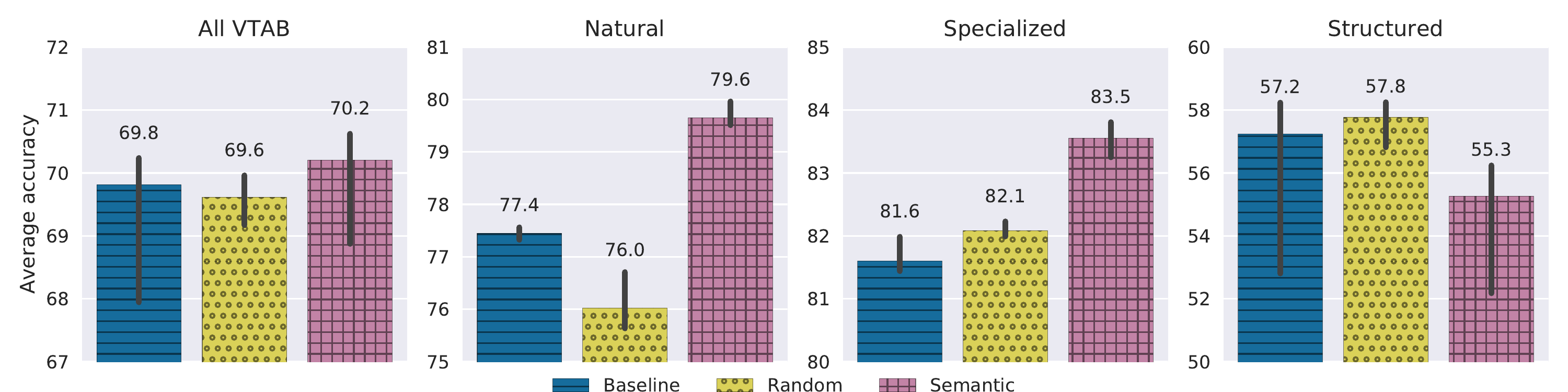}}
\caption{Results on VTAB with 1\,000 examples per dataset achieved by experts 
trained on random subsets of \JFT and experts trained on semantically meaningful
subsets. For each dataset in VTAB, the median accuracy over 30 trials is 
considered.  The results of the datasets in each group are averaged. 
The error bars show the (percentile) bootstrapped confidence intervals at 
95\% level.}
\label{fig:vtab_test_expert_vs_many_models}
\end{figure}

In principle, it was not even clear if this approach with random experts would outperform the baseline.
\Cref{fig:vtab_test_expert_vs_many_models} (a) and (b) show our results for adapter and full experts respectively.

In both cases, the overall performance drops when we replace semantic experts with random ones.
This difference seems stronger in the case of adapter-based experts.
However, notice that the full expert results are very influenced by 
strong negative results in one of the structured datasets 
(Clevr Count), as shown in \cref{tab:vtab_per_dataset_all}.
Also, random experts results are comparable to the baseline.
Note that random experts are not dumb; they are just a diverse set of models with the general flavor of the upstream dataset.
The algorithm can still benefit from their diversity when confronted 
with a new task, whereas we expect them to be more similar to each other than in the semantic case.

The semantic experts are trained mostly on natural slices, and we see a large improvement in the \textsc{Natural} tasks when we use them (2.7\% and 4.7\% gains for adapters and full, respectively, compared to random experts).
This reinforces the idea that experts in the right domain can be very helpful.
Moreover, in \textsc{Natural} tasks, the baseline outperforms random experts; this suggests that here more data is better unless data is smartly selected.

As we have hypothesized before, it seems that our natural-image experts do not provide a meaningful expertise or competitive edge on \textsc{Structured} tasks.
We see a large improvement on \textsc{Specialized} tasks when using full experts, while the effect is not there for adapters.
Accordingly, we would not read too much into these results.

\end{document}